\documentclass[journal,twoside,web]{ieeecolor}
\usepackage{generic}
\usepackage{amsmath,amssymb,amsfonts}
\usepackage{mathtools}
\usepackage{algorithm,algorithmic}
\usepackage{graphicx}
\usepackage{subcaption}
\usepackage{hyperref}
\usepackage[cmyk,dvipsnames]{xcolor}
\usepackage{soul}
\usepackage{nameref}
\usepackage{tabulary}
\usepackage{booktabs}
\usepackage[backend=biber,style=ieee,natbib=true]{biblatex} %added
\hypersetup{hidelinks=true}
\usepackage{textcomp}
\usepackage{siunitx}
\usepackage{amsmath} % Placing argument below min/max

\DeclareMathOperator{\diag}{diag}
\DeclareMathOperator{\abs}{abs}
\DeclareMathOperator{\len}{len}

\DeclareMathOperator{\mean}{mean}
\DeclareMathOperator{\SD}{SD}
\DeclareMathOperator{\BatchProcessEMG}{\textsc{BatchProcEMG}}
\DeclareMathOperator{\PeakCrossCorr}{\textsc{PeakCrossCorr}}
\DeclareMathOperator{\OptimizationProblem}{\textsc{OptimProb}}

\def\BibTeX{{\rm B\kern-.05em{\sc i\kern-.025em b}\kern-.08em
    T\kern-.1667em\lower.7ex\hbox{E}\kern-.125emX}}
\begin{document}
\title{Koopman-driven Grip Force Prediction Through EMG Sensing}
\author{Tomislav Bazina, Ervin Kamenar, Maria Fonoberova, Igor Mezić, \IEEEmembership{Fellow, IEEE}
\thanks{%Manuscript received Xxxxx xx, 2024; revised Xxxxx xx,
%2024; accepted Xxxxxx xx, 2024. Date of publication Xxxxxx xx,
%2024; date of current version Xxxxxx x, 2024.
This work was supported by the U.S. Air Force Office of Scientific Research (AFOSR) under Award FA9550-22-1-0531 and by the University of Rijeka under Grant uniri-iskusni-tehnic-23-47.
\textit{(Corresponding author: Ervin Kamenar).}}
\thanks{Tomislav Bazina is with the University of Rijeka, Faculty of
Engineering, 51000 Rijeka, Vukovarska 58, Croatia (e-mail:
tomislav.bazina@uniri.hr).}
\thanks{Ervin Kamenar is with the University of Rijeka, Faculty of Engineering,
51000 Rijeka, Vukovarska 58, Croatia (e-mail: ekamenar@uniri.hr).}
\thanks{Igor Mezić is with the University of California, Santa Barbara, CA,
93101 USA (e-mail: mezic@ucsb.edu).}
\thanks{The authors are with AIMdyn, Inc., Santa Barbara, CA, 93101, USA (e-mail:
tomislav@aimdyn.com; mezici@aimdyn.com; mfonoberova@aimdyn.com;
ervin@aimdyn.com).}}

\maketitle

\begin{abstract}
Loss of hand function due to conditions like stroke or multiple sclerosis impacts daily activities. Robotic rehabilitation provides tools to restore hand function, while surface electromyography (sEMG) enables the adaptation of the device's force output to the user’s condition, thus enhancing rehabilitation outcomes. This study focuses on accurately predicting grip force during medium wrap grasps using a single sEMG sensor pair, addressing the challenge of escalating sensor requirements. We conducted sEMG measurements on 13 subjects at two forearm positions, validating results with a hand dynamometer. 
Established flexible signal-processing steps achieved high peak cross-correlations between the processed sEMG signal and grip force. Influential parameters were subsequently identified through sensitivity analysis. Leveraging a novel data-driven Koopman-based approach and problem-specific data lifting, we devised a method for the estimation and short-term prediction of grip force from processed sEMG signals. The method achieved a weighted mean absolute percentage error (wMAPE) of \(\sim\)5.5\% for grip force estimation and \(\sim\)17.9\% for 0.5-second predictions. The methodology proved robust regarding precise electrode positioning, as the effect of sensing position on error metrics was non-significant. The algorithm executes exceptionally fast, processing, estimating, and predicting a 0.5-second sEMG signal batch in just \(\sim\)30 ms, facilitating real-time implementation.
\end{abstract}


\section{Introduction}
\label{sec:introduction}
\IEEEPARstart{T}{he} electromyography (EMG)-based robotic rehabilitation outperforms conventional methods, such as constraint-induced movement and physical therapy. It improves motor recovery, reduces spasticity, and enhances patient engagement by maximizing their voluntary action \cite{huo2023effects}. Estimating and predicting grip force from real-time EMG signals enables accessing the control variable---force---allowing robotic assistance to supplement patients' voluntary effort adaptively.

The non-invasive sEMG technique measures muscle activity by detecting electrical signals generated by motor units (MU), which are activated by motor neurons. The EMG signal is obtained as an interference of firing signals from each MU, known as motor unit activation potential (MUAP). Factors such as electrode configuration (placement, size, inter-distance), tissue properties (fat layers, skin conductivity), temporal/spectral firing patterns, and cross-talk from neighboring muscles affect signal variability \cite{Stegeman2000}.

In \cite{Ma2020}, pinching force was predicted using a 6-channel sEMG sleeve with RMS features and the gene expression programming algorithm, achieving RMSE errors of 7.5--8.5\% and cross-correlation coefficients up to 95\%. The study examined four MVC levels (20\%, 40\%, 60\%, and 80\%) but excluded predictions near the transient states between these levels. In \cite{Khan2024}, sEMG with four bipolar electrodes and finger force signals were used to predict grip force with 90\% accuracy during the transient phase, utilizing five EMG and eleven force features. Optimal sensing required 2--4 features across three positions. Another study \cite{Baranski2014} identified the brachioradialis muscle as optimal for grip force sensing based on EMG signal strength. In \cite{Siavashani2023}, EMG signals were acquired using an eight-channel Myo's armband, and an LSTM network predicted normalized pinching force 1, 3, and 5 seconds ahead directly from the EMG data. Based on \cite{Zhang2021}, which used eight sEMG sensors and 24 healthy subjects, extrinsic muscle coordination reliably predicted grip and pinch force levels due to its greater sensitivity to force changes compared to intrinsic muscles. 

Some studies, including \cite{Martinez2020a} and \cite{Martinez2020}, have explored predicting gripping force from the transient phase of EMG signals, which captures the initial burst of muscle activity. In \cite{Martinez2020a}, a 192-channel sEMG setup was used with \num{12} participants to predict grasp force, achieving absolute errors as low as 2.5\% of MVC using ten features and regularized linear regression. Meanwhile, \cite{Martinez2020}, utilized 8 sEMG sensors placed on the forearms of 16 participants. The model used ten EMG features to predict gripping force 330 ms ahead with the elastic net regression, resulting in errors of 2\% of MVC.

The analysis above, along with findings in \cite{Wu2021}, show that while the majority of the research has achieved favorable results in terms of accuracy, it relies on a high—and constantly increasing—number of necessary sEMG sensors for the accurate prediction of grasping force. Moreover, forecasts during the transient state are rarely considered. Additionally, existing methods often overlook advanced signal processing for extracting meaningful information from noisy EMG signals, which is a crucial step in causal grip force modeling. While Koopman operator theory (KOT) \cite{Mezic2005} has demonstrated success in rehabilitation (e.g., Functional Electrical Stimulation \cite{singh2025koopman}) and multi-modal physiological dynamics reconstruction via deep Koopman auto-encoders \cite{qian2022deep}, we present the first online framework for grip force estimation and short-term forecasting using a single sEMG sensor pair.
%While Koopman operator theory (KOT) \cite{Mezic2005} has seen successful rehabilitation applications (e.g., Functional Electrical Stimulation \cite{singh2025koopman}), we present the first online framework for grip force estimation and short-term forecasting. %Our research is the first to develop an online modeling and signal-processing framework for estimating and short-term forecasting grip force based on Koopman operator theory (KOT) \cite{Mezic2005}.
The main contributions of this work are:
\begin{itemize}    
    \item We devised and optimized a composition of EMG signal processing methods, achieving high peak cross-correlations between EMG and grip force signals using a single sensing position on the forearm.
    \item We devised a novel Koopman-based data-driven approach with problem-specific observables for estimating and short-term predicting grip force in real-time during both transient and plateau phases.
    \item The methodology enables fast execution, thus facilitating real-time implementation.
\end{itemize}

\section{Materials and methods}
\label{sec:materials_and_methods}

The first part of the study involves collecting time-series data on human\footnote{Informed consent was obtained from all human subjects involved. The research was performed under the oversight of the University of Rijeka Faculty of Engineering Ethics Committee, reference number 2409.17340.} muscle activity through sEMG and measuring hand-grip force with a hand dynamometer. The dynamometer's shape necessitated a medium wrap type of grasp, following the taxonomy developed by \cite{Cutkosky1989} and refined by \cite{Feix2016}. To minimize prediction errors, we considered sEMG positioning on the forearm to maximize signal measurement. Advanced signal processing was applied to extract relevant features from the EMG signals and establish a robust correlation with grip force. Offline optimization of decision variables was also conducted.

We use KOT to represent nonlinear dynamics with linear operators, distinguishing between "dynamic" and "static" types \cite{Mezic2021}. Both act on observables---functions mapping the system's state to scalar or vector values---rather than directly on the state. The dynamic operator advances observables in time, making it suitable for short-term grip force forecasting, while the static operator maps observables between different spaces, estimating current grip force from EMG signals.

An important step is the signal filtering, essential for proper inferences using EMG signals, which are often affected by noise and cross-talk between muscles. The signals are typically processed using notch filtering at 50 Hz to eliminate ground noise \cite{Khan2024}, bandpass (BP) filtering between 10 and 500 Hz \cite{Martinez2020,Martinez2020a,Zhang2021,Wu2021,Baranski2014}, and BP filtering that captures only power spectrum peaks between 20 and 60 Hz \cite{Siavashani2023}. In contrast, we focused on extracting meaningful features (comparable to observables in KOT) from the sEMG signal---those highly correlated with grip force---that can be applied within the KOT framework. This strategy aligns with the structural approach to sEMG modeling in \cite{Stegeman2000}, which states that recruiting MUs is fundamental for generating muscle force. Greater MU recruitment and higher firing rates amplify force output, with firing rates increasing almost linearly with muscle force. These firing patterns and the interference between the active MUs influence the characteristics of the measured sEMG. We aim to isolate the difficult-to-identify signal components---specifically the recruited MUs, their interference patterns, and firing rates---that predominantly contribute to measured grip force during a medium wrap, to separate meaningful signals from unwanted noise. To achieve this, we utilized Fast Fourier Transformation (FFT) and sensitivity analysis (SA) to develop a spectral mask that selectively targets specific spectral components. 

The ultimate goal is to develop a fully integrated module that accurately estimates and predicts exerted grip force from EMG signals using KOT, encompassing advanced signal processing and personalized calibration protocols. 

% Section \ref{sec:experiment} outlines our experimental design, Section \ref{sec:sensitivity_analysis} details EMG signal optimization, and Section \ref{sec:koopman} examines KOT implementation for grip force estimation and prediction.

\subsection{Experimental Design}
\label{sec:experiment}
Two complementary wireless sensing devices were utilized: an EMG sensing unit and a grip dynamometer. The EMG sensing unit records electrical activity on the skin surface associated with muscle contractions, while the dynamometer measures gripping force. The EMG signals were sampled at \qty{1}{\kHz}, while the dynamometer data was acquired at \qty{200}{\Hz}. Detailed device specifications and acquisition parameters are provided in Appendix~\ref{app:hardware}. The reference dynamometer was calibrated as described in Appendix~\ref{app:calibration}.

\begin{figure}[!t]
    \centering
    \begin{subfigure}{0.49\columnwidth}
        \includegraphics[width=1\textwidth]{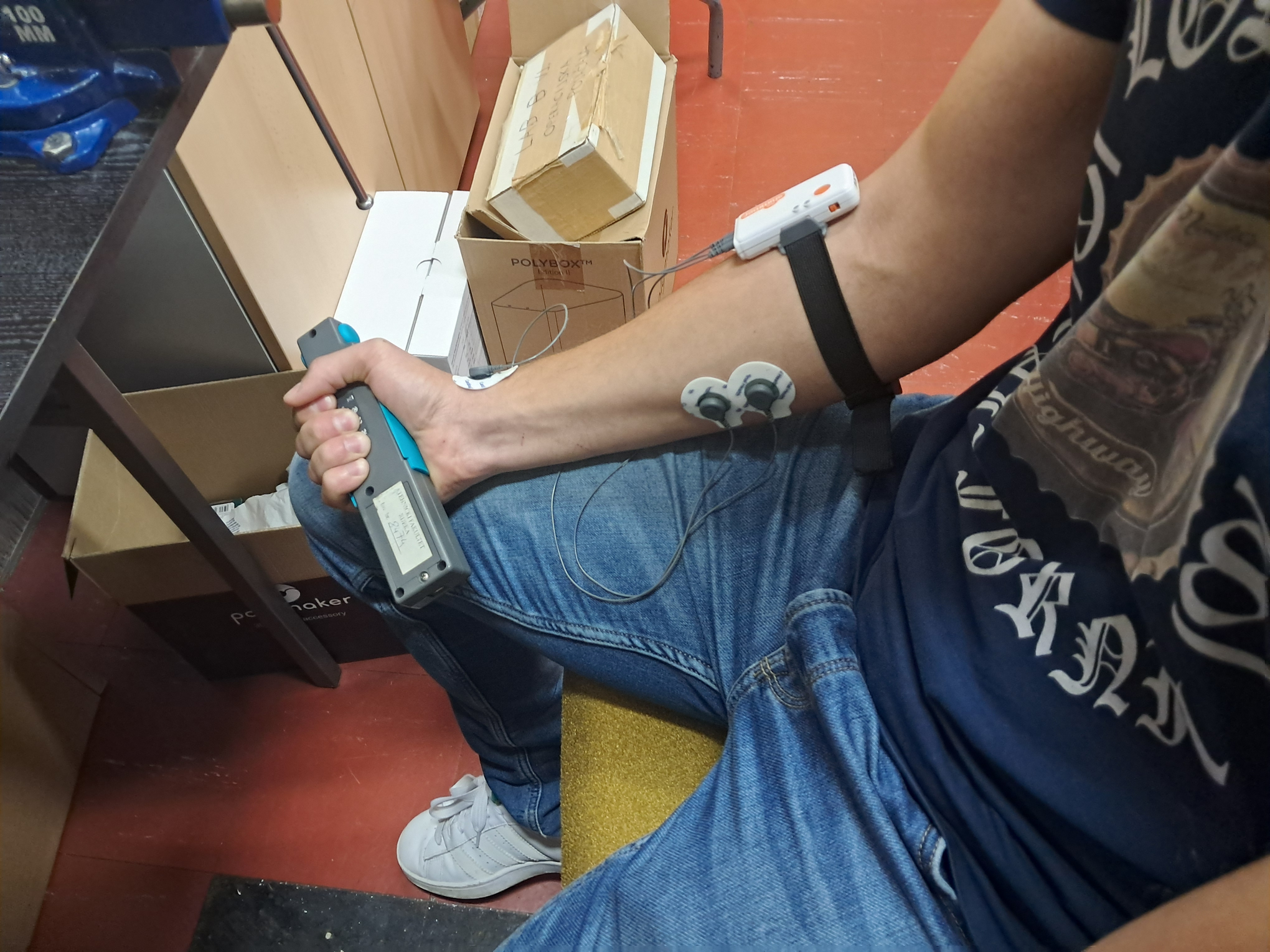}
        % \caption{First position} \label{fig:emg-position-1}
        \caption{} \label{fig:emg-position-1}
    \end{subfigure}
    \begin{subfigure}{0.49\columnwidth}
        \includegraphics[width=1\textwidth]{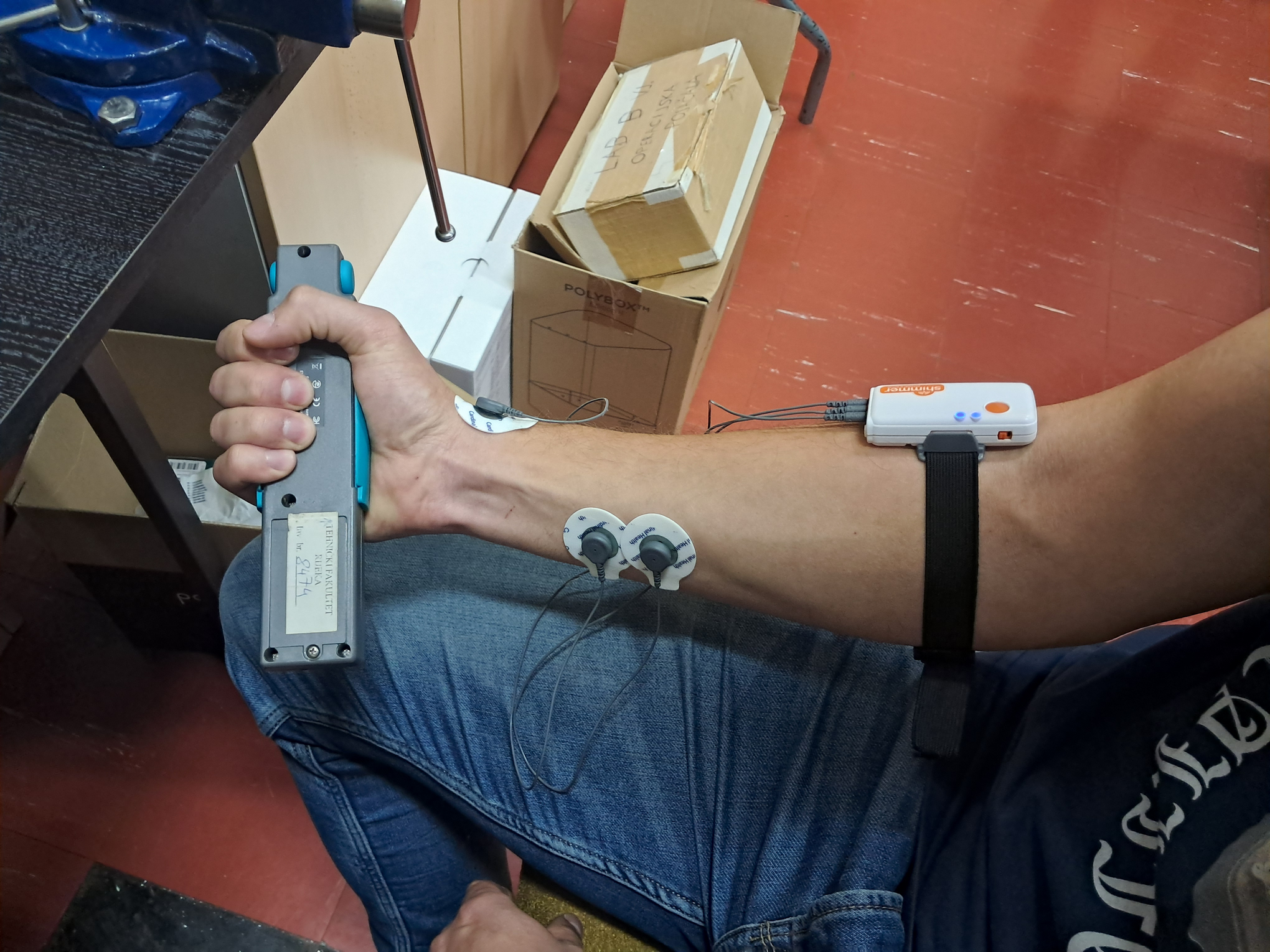}
        % \caption{Second position} \label{fig:emg-position-2}
        \caption{} \label{fig:emg-position-2}
    \end{subfigure}
    \caption{Placement of sEMG electrodes near flexor carpi ulnaris muscle on the forearm: (a) Position 1 (P1); (b) Position 2 (P2).}
    \label{fig:emg-position}
\end{figure}

Various electrode positions were tested, and two forearm positions (see Fig.~\ref{fig:emg-position}) were selected for further analysis based on signal strength observed in a simple screening experiment. A two-factor randomized block design (RBD) \cite{Heckert2012} was chosen, as described in Table~\ref{tab:rbd-design}.  Electrode positioning followed configurations from prior studies \cite{Baranski2014,Wu2021}. The experiment involved \num{13} male participants (ages \numrange{22}{24}), with subject variability controlled through blocking to isolate measurement position effects on estimation and prediction errors. Randomization was implemented within blocks. Skin preparation included hair removal and alcohol cleansing at measurement positions. Participants were given time to familiarize themselves with the required force levels. For all trials, the initial \qty{5}{\s} were used to zero the dynamometer signal at each position.
% Reduce spacing between columns in a table
\setlength{\tabcolsep}{3.5pt}
\begin{table}[!h]
\caption{Summary of two-factor RBD experiment.}
\label{tab:rbd-design}
\centering
\begin{tabulary}{\columnwidth}{CCCCC}
\toprule
\textbf{Subject Levels} & \textbf{Position levels}  & \textbf{Grip force levels [\%]} & \textbf{Replications} & \textbf{Runs \quad} \\
\midrule
ac, dp, ds, js, lb, lk, lm, ln, md, mm, nk, pb, ss & 1, 2 & 100, 75, 50, 25, 0 & 2 & 52 \\ 
\bottomrule
\end{tabulary}
\end{table}

\subsection{Procedural Parameter Optimization and sEMG Procesing Steps}
\label{sec:sensitivity_analysis}
The optimization procedure maximizes the mean peak cross-correlation between processed EMG signal \(\textbf{e}_\textrm{proc}\) and the synchronously collected grip force \(\textbf{g}\). This metric is selected for its effectiveness in capturing signal similarity across time lags, which vary between subjects and measurements. 
Algorithm \(\PeakCrossCorr\) (see Algorithm~\ref{alg:cross_correlate_emg_grip} in Appendix~\ref{app:peak_cross_correlate_emg_grip}) computes the per-run peak cross-correlation. Multi-step SA and optimization identify key spectral components and narrow decision vector bounds near the optimum using scatterplot projections with smoothed trends and means, generated via Latin hypercube (LH) sampling. This integrated approach ensures robust cross-correlations, consistent performance, and stability by avoiding unstable maxima despite small variations.

\begin{figure}[!t]
    \centering
    \includegraphics[width=\columnwidth]{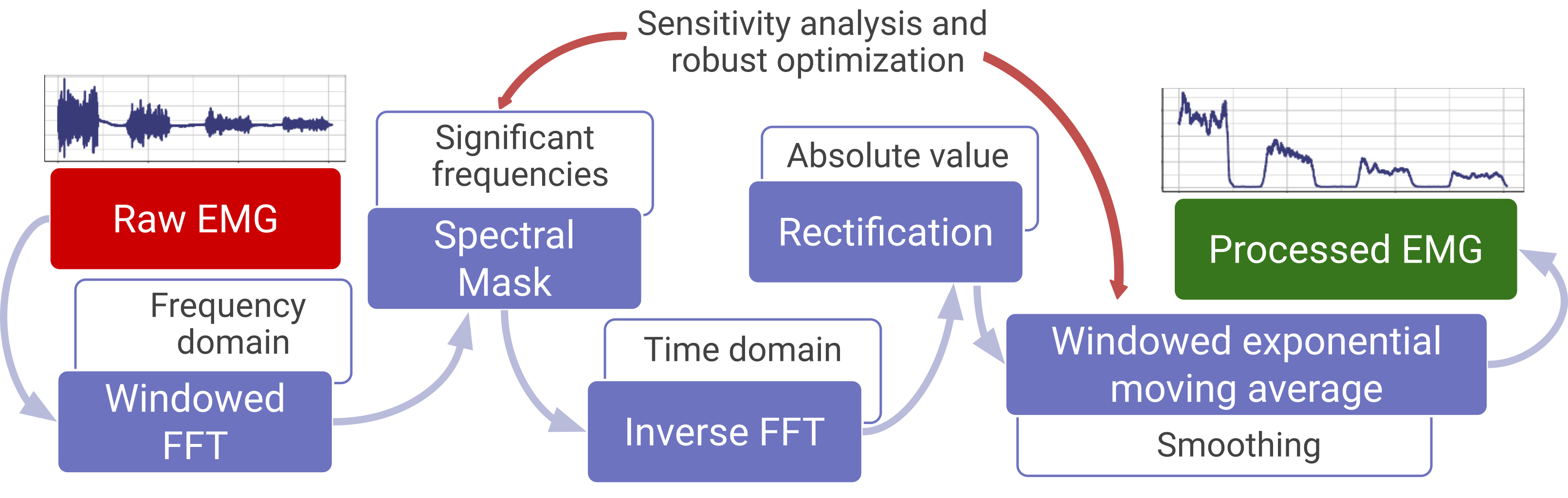}
    \caption{Signal processing steps for obtaining optimal processed EMG regarding cross-correlation with measured grip strength.}
    \label{fig:signal-processing-steps}
\end{figure}

First, EMG signals are processed in mini-batches of approximately \num{0.5} seconds (\num{496} data points) with a sampling rate of \qty{992.97}{\Hz}. This batch size enables quasi-stationary analysis while maintaining sensitivity to physiologically relevant variations in grip force. This aligns with \citet{Miklashevsky2022}, who demonstrated that while initial stimulus processing activates the motor system early (\qtyrange{100}{130}{\ms}), complex decision-making processes are reflected in grip force patterns only after \qtyrange{300}{350}{\ms}. The selected batch size optimizes the tradeoff between frequency and temporal resolution. It is sufficient for reliable frequency analysis via Short-Time Fourier Transform (STFT) and capturing meaningful changes in signal dynamics. We transform each batch using FFT (resolution: \qty{2.002}{\Hz}), followed by the application of an optimized spectral mask \(\mathbf{w}_\text{mask}\) to modify frequency components selectively. After inverse FFT, the signal is rectified by taking absolute value and smoothed using an exponential moving average (MA) with optimized window size \(w_\text{EMA}\) and decay factor \(\alpha_\text{EMA}\). The exponential MA window \(\mathbf{wnd}_{\text{EMA}}\) includes previous batch data for the first \(w_\text{EMA}-1\) points to ensure smooth transitions between batches. The batch processing pipeline for optimal EMG signal extraction is illustrated in Fig.~\ref{fig:signal-processing-steps}, with the detailed  Algorithm~\ref{alg:processing_emg}: \(\BatchProcessEMG\) provided in Appendix~\ref{app:batch_process_emg}.

The high-dimensional optimization problem was formulated to maximize the mean peak cross-correlation \(\textbf{r}_\text{peak,all}\) across all 52 experiment runs using a decision vector comprised of 250 variables. The entries and initial bounds of the decision vector:
\begin{itemize}
    \item Spectral mask \(\mathbf{w}_\text{mask}\) with \num{248} entries corresponding to each frequency bin (without DC offset) in range \numrange{0}{5},
    \item Exponential MA window size \(w_\text{EMA}\) in range \numrange{2}{495},
    \item Exponential MA decay factor \(\alpha_\text{EMA}\) in range \numrange{0}{0.05}.
\end{itemize}
The optimal decision vector is obtained by maximizing the objective function:
\begin{equation}
\begin{aligned}
    \underset{\textbf{w}_\textrm{mask}, w_\textrm{EMA}, \alpha_\textrm{EMA}}{\arg\max} \OptimizationProblem \big( \textbf{w}_\textrm{mask}, w_\text{EMA}, \alpha_\text{EMA} \big),
\end{aligned}
\end{equation}
where the \(\OptimizationProblem\) is described in Algorithm~\ref{alg:optimization_problem}. The previously described algorithms for EMG batch processing and computing peak cross-correlation are applied to each of the 52 experimental runs, extracting paired EMG and grip force data \((\mathbf{e}_\textrm{raw}, \mathbf{g})\) from the complete dataset \((\mathbf{e}_\textrm{raw,all}, \mathbf{g}_{\text{all}})\).
\begin{algorithm}[!h]
\caption{\( \OptimizationProblem \big( \textbf{w}_\textrm{mask}, w_\text{EMA}, \alpha_\text{EMA} \big) \)}
\label{alg:optimization_problem}
\algsetup{indent=0.8em, linenosize=\scriptsize}
\begin{algorithmic}
\STATE \(\textbf{r}_\text{peak,all} \leftarrow \left(\,\right)\) \COMMENT{Initialize empty tuple}
\FOR{each run \( \big( \textbf{e}_\textrm{raw}, \textbf{g} \big) \) in all \( \big( \textbf{e}_\textrm{raw,all}, \textbf{g}_{\text{all}} \big) \)}
\STATE \( \textbf{e}_\text{proc} \leftarrow \BatchProcessEMG \big( \textbf{e}_{\textrm{raw}}, \textbf{w}_\textrm{mask}, w_\text{EMA}, \alpha_\text{EMA} \big) \)
\STATE \( \textbf{r}_\text{peak} \leftarrow \PeakCrossCorr \big( \textbf{e}_{\textrm{proc}}, \textbf{g} \big) \)
\STATE \(\textbf{r}_{\text{peak,all}} \leftarrow \big(\textbf{r}_{\text{peak,all}}, \textbf{r}_{\text{peak}}\big)\) \COMMENT{Concatenate objective}
\ENDFOR
\RETURN \( \mean \big( \textbf{r}_\text{peak,all} \big) \)
% \RETURN \( \arg\max\limits_{\textbf{w}_\textrm{mask}, w_\textrm{EMA}, \alpha_\textrm{EMA}} \big( \mean \big( \textbf{r}_\text{peak,all} \big) \big) \)
\end{algorithmic}
\end{algorithm}

The DC offset was set to zero and excluded from the optimization process. Initial mask bounds allowed either complete removal of a spectral component's amplitude or amplification by up to five times. In contrast, the decay factor bounds were designed to support simple MA (if set to 0) and exponential MA (if greater than 0). Broad bounds on the window size also enabled fine-tuning of the applied smoothing.

Due to differences in sampling rates---where the EMG signal is sampled \num{\sim 5} times faster than the dynamometer---the grip force signal \(\textbf{g}\) was resampled using EMG timestamps. Intermediate points were estimated using linear regression to ensure proper signal alignment for cross-correlation computation. We hypothesize that not all \num{248} FFT frequency bins significantly impact the cross-correlation. To address this, SA is performed in conjunction with optimization to narrow the problem's scope and identify the most influential spectral components on the optimization problem in Algorithm~\ref{alg:optimization_problem}.
A preliminary Sobol SA is conducted separately for positions P1 and P2, utilizing grouped spectral mask variables to provide a high-level understanding of how cross-correlation sensitivity is affected by smoothing and filtering. Details regarding the preliminary SA procedure, parameters, and sample size are outlined in Appendix~\ref{app:preliminary_sa}.

\begin{figure*}[!t]
    \centering
    \begin{subfigure}{0.33\textwidth}
        \includegraphics[width=1\textwidth]{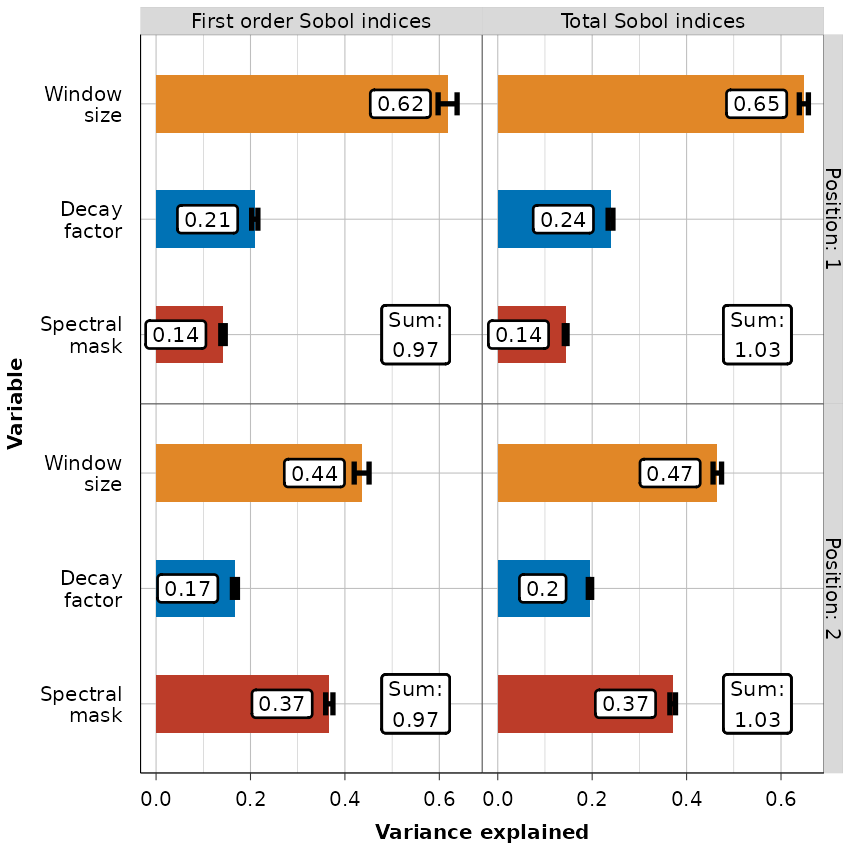}
        \caption{} \label{fig:preliminary-sensitivity-sobol}
    \end{subfigure}
    \begin{subfigure}{0.66\textwidth}
        \includegraphics[width=1\textwidth]{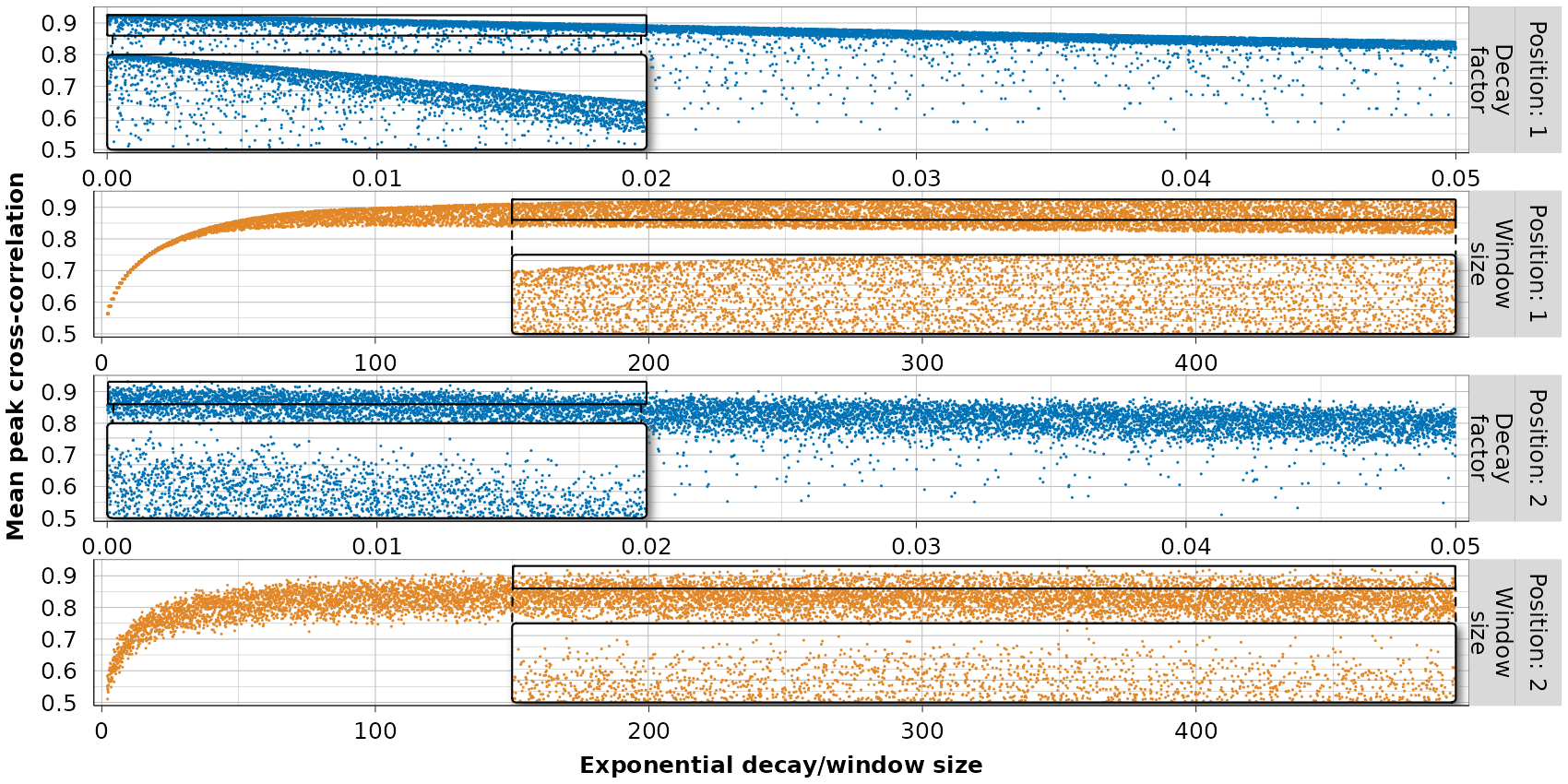}
        \caption{} \label{fig:preliminary-sensitivity-latin}
    \end{subfigure}
    \caption{Preliminary sensitivity analysis for two sensing positions: (a) first-order and total-order sensitivity indices, and (b) LH sampling projections onto the decay factor and window size variables.}
    \label{fig:preliminary-sensitivity}
\end{figure*}

Fig.~\ref{fig:preliminary-sensitivity-sobol} presents the obtained first-order and total-order sensitivity indices (SIs) and their sums. SIs sums are \num{\sim 1}, indicating that they can reliably approximate the percentage of output variance attributable to each variable, and interactions between variables can be safely ignored. When sampling the problem within the initial bounds, the contributions of smoothing parameters and the spectral mask to the variance in mean peak cross-correlation depend on the measurement position. Smoothing window size \(w_\text{EMA}\) is the most influential factor, explaining the majority of output variance across both positions. The decay factor contributes \qty{\sim 20}{\percent}, while the spectral mask contribution varies most: \qty{14}{\percent} on P1 to \qty{37}{\percent} on P2.
Additional LH sampling with \num{10000} decision vector samples was conducted within the same bounds to investigate the partial contributions of smoothing parameters further. Fig.~\ref{fig:preliminary-sensitivity-latin} displays the resulting projections onto the decay factor \(\alpha_\text{EMA}\) and window size \(w_\text{EMA}\). Increasing \(w_\text{EMA}\) improves the peak cross-correlation until it reaches a plateau, supporting the decision to narrow the window size bounds to the \numrange{200}{495} range for subsequent steps. Conversely, increasing \(\alpha_\text{EMA}\) shows a decreasing trend in peak cross-correlation, so we narrowed the range to \numrange{0}{0.01}. The grouped SA was repeated with the narrowed parameter bounds, where the spectral mask contributed to \qty{88}{\percent} of the variance in P1 and \qty{95}{\percent} in P2. This highlights the necessity of ungrouping the mask for more detailed analysis in the subsequent steps.

\begin{figure*}[!t]
    \centering
    \includegraphics[width=\textwidth]{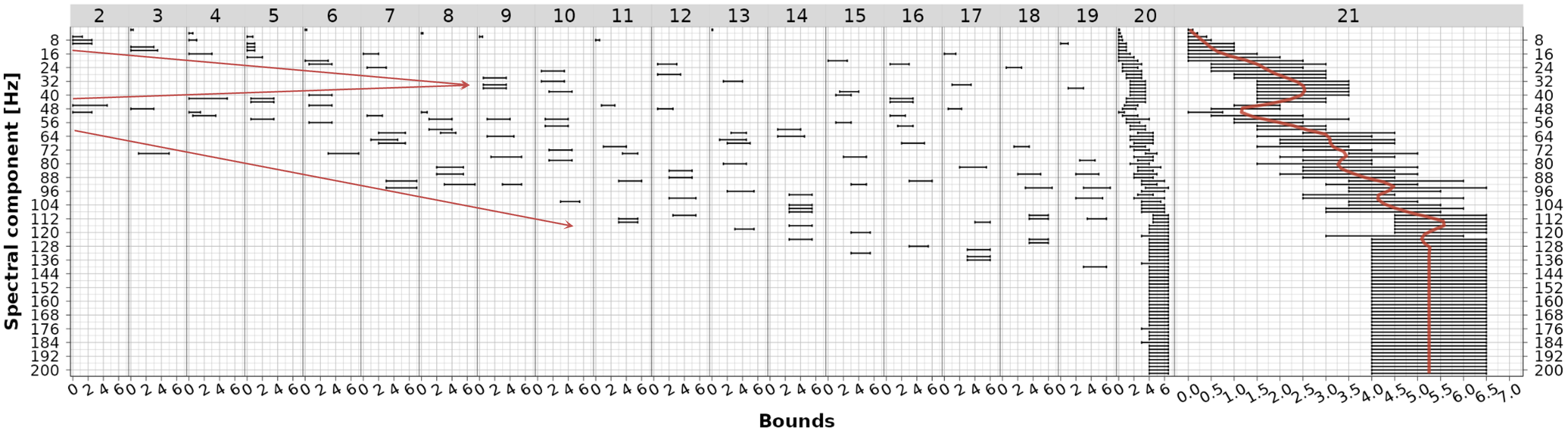}
    \caption{Iterative multi-step sensitivity analysis (steps \(2^\text{nd}–21^\text{st}\)) and simultaneous optimization by progressively narrowing the decision vector bounds and identifying the most influential frequency components. Initial bounds were set in the range \numrange{0}{5}.}
    \label{fig:frequency-bounds-steps}
\end{figure*}

For iterative, multi-step, ungrouped SA, a more efficient RBD-FAST SA method \cite{Tarantola2006} was used. Details regarding the simultaneous SA and optimization procedure steps are outlined in Appendix~\ref{app:multistep_sa}. The top spectral components with narrowed bounds after each SA and optimization step are presented in Fig.~\ref{fig:frequency-bounds-steps}. In a step-by-step analysis, a reverse funnel-shaped pattern can be observed (indicated by the two red arrows), with one starting point at low-frequency components and the other around \qty{50}{\Hz}. This pattern highlights the transition from the most influential spectral components to the less influential ones.
After the \(19^\text{th}\) step of ungrouped SA, the cumulative contribution of the SIs corresponding to spectral mask for frequencies \qty{>= 204}{\Hz} amounted to less than \qty{2}{\percent}. Therefore, we reduced the spectral mask \(\textbf{w}_\textrm{mask}\) to \num{101} components (\qtyrange{2}{202}{\Hz}, step \qty{\sim 2}{\Hz}), consistent with \cite{Konrad2006}, which shows most sEMG power is below \qty{250}{\Hz}. Last two SA and optimization steps (\(20^\text{th}, 21^\text{st}\)) reveal interesting trends:
\begin{itemize}
    \item Increasing decay factor \(\alpha_\text{EMA}\) decreases cross-correlation, and optimal values are close to zero (\numrange{0}{5e-4}), indicating a need for simple MA.
    \item Window size \(w_\text{EMA}\) optimum is in the \numrange{275}{330} range.
\end{itemize}
The description of the optimal spectral mask is detailed in Section~\ref{sec:results}. 
With the optimal set of signal processing procedures and parameters, next section outlines the Koopman-driven methodology for estimating current and predicting future values of the grip force from the processed EMG signal.

\subsection{Koopman-driven Estimation and Prediction of Hand Grip Force} \label{sec:koopman}
The framework begins with a one-time \qtyrange{20}{30}{\s} calibration experiment using sEMG sensors and a dynamometer, with grip force levels as in Table~\ref{tab:rbd-design}. The processed EMG signal \(\textbf{e}_{\text{proc}}\) is used to train the estimation model. Data is processed batch-by-batch for real-time application to estimate current grip force \(\textbf{g}\) under the window. This estimation is used to train a prediction model that forecasts grip force over a \qty{0.5}{\s} horizon.

\subsubsection{Koopman Operator Theory}
\label{subsec:KOT}
The state-space representation of a dynamical system involves an \(n\)-dimensional manifold \(\mathcal{M}\), where states \(\mathbf{x}\) evolve discretely over time:
\begin{equation}
    \begin{aligned}
    \mathbf{x}[i+1] = \mathbf{F}\big(\mathbf{x}[i]\big),
    \end{aligned}
    \label{eq:dynamicalMap}
\end{equation}
where \(\mathbf{F}\mathbin{:} \mathcal{M} \to \mathcal{M}\) is the potentially nonlinear state transition function, and \(\mathbf{x}[i+1]\) represents the time-shifted state. To address the challenges of modeling complex nonlinear dynamics, we adopt an operator-theoretic perspective on the dynamics of observables \cite{Mezic2005}. The nonlinear system (\ref{eq:dynamicalMap}) is mapped onto observable dynamics \(\phi(\mathbf{x})\), typically complex-valued, but here we focus on real-valued \(\phi\mathbin{:} \mathcal{M} \to \mathbb{R}\). 
Collecting all possible observables constitutes a vector space that is generally infinite-dimensional. The Koopman operator \(\mathcal{K}\), describing observable evolution over \(\Delta t\), is defined as \cite{Schmid2022}:
\begin{equation}
    \begin{aligned}
        \phi\big(\mathbf{x}[i+1]\big) = \mathcal{K}\phi\big(\mathbf{x}[i]\big) = \phi\Big(\mathbf{F}\big(\mathbf{x}[i]\big)\Big).
    \end{aligned}
    \label{eq:dynamical_evolution}
\end{equation}
$\mathcal{K}$ remains linear even for nonlinear underlying system \cite{Mezic2005}.

\subsubsection{Estimating Grip Force}
\label{subsec:estimation}
\begin{figure}[!t]
    \centering
    \includegraphics[width=\columnwidth]{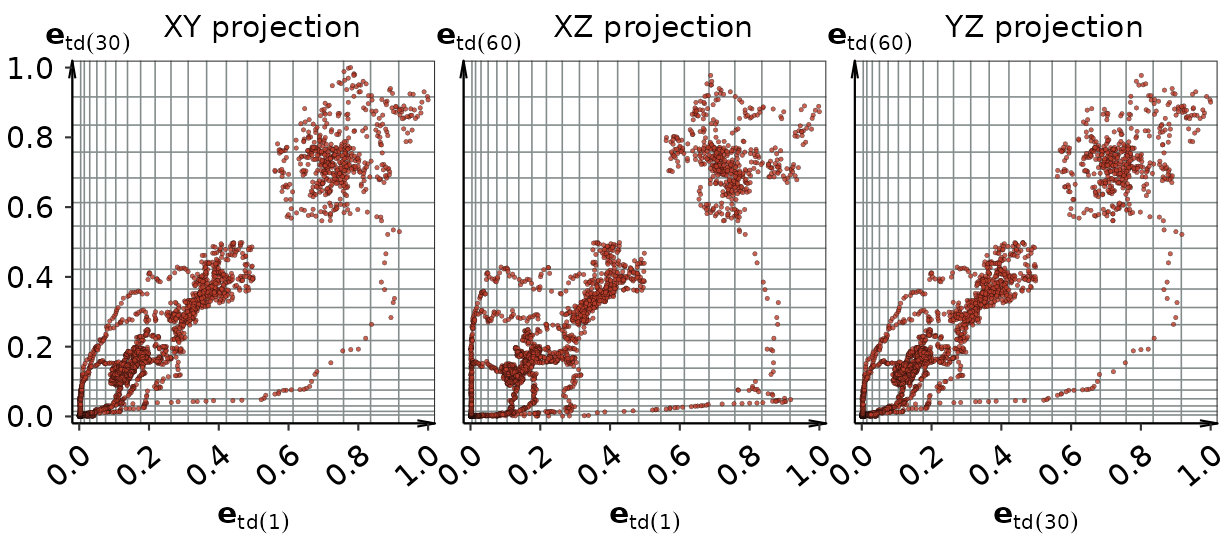}
    \caption{Example projections of 3D gridded indicator observables \(\mathbf{e}_{\mathrm{I},S_{ijk},29,59}\) with optimal grid division.}
    \label{fig:gridded-indicator-observables}
\end{figure}

The Koopman operator can also describe ``static`` nonlinear maps between different spaces \(\mathcal{M} \to \mathcal{N}\) \cite{Mezic2021}, a property we leverage to estimate current-batch grip force from processed EMG. Through lifting and proper choice of observables, we can describe static nonlinear maps using spaces of observables and a linear mapping operator \(\mathcal{K}_\text{e}\mathbin{:} \mathcal{O}_\mathcal{M} \to \mathcal{O}_\mathcal{N}\). The processed EMG signal \(\textbf{e}_\text{proc}\) and measured grip force \(\textbf{g}\) are lifted using yet-to-be-determined vectors of functions \(\mathbf{\phi}\) and \(\psi\), respectively. The input and output matrices of the lifted variables, where \(\big(\textbf{e}_\text{proc}[i], \textbf{g}[i]\big)\) represents a single data pair realization, can be expressed as:
\begin{equation}
\begin{aligned}
    E = \phi\big(\textbf{e}_\text{proc}\big), \quad 
    G = \psi\big(\textbf{g}\big).
\end{aligned}
\end{equation}
The approximation of the static Koopman operator is obtained by minimizing the Frobenius norm \cite{Mezic2021}:
\begin{equation}
    \begin{aligned}
    \min_{\mathcal{K}_\text{e}} \| G - \mathcal{K}_\text{e}E \|_F \quad \to \quad \overline{\mathcal{K}}_\text{e} = GE^{\dagger},
    \end{aligned}
    \label{eqn:static-koopman}
\end{equation}
where $^{\dagger}$ is the SVD-based Moore-Penrose pseudoinverse. The estimation Koopman operator \(\overline{\mathcal{K}}_\text{e}\), trained on the complete personalized calibration experiment, is applied sequentially to each \qty{\sim 0.5}{\s} batch window to generate real-time grip force estimations. Empirical analysis showed 8-fold downsampling (\(993 \to \qty{124}{\Hz}\)) of processed EMG (Fig.~\ref{fig:proc-est-signal}) and normalization to \([0, 1]\) significantly reduced grip force estimation error. We employed Hankel lifting with time-delay embedding, a technique well-suited for capturing the system's temporal dynamics \cite{Frame2023}. The Hankel data matrix \(E\), constructed from \(\textbf{e}_\text{proc}\) with \(N\) data points, incorporates state \(\mathbf{e}_0\) in the first row and \(d\) time-delayed observables \(\mathbf{e}_{\textrm{td}(i)}\) in subsequent rows:
\begin{equation}
    \begin{aligned}
    \resizebox{0.91\columnwidth}{!}{\(
    E = 
    \begin{bmatrix}
    \mathbf{e}_0 \\
    \mathbf{e}_{\textrm{td}(1)} \\
    \vdots \\
    \mathbf{e}_{\textrm{td}(d-1)} \\
    \mathbf{e}_{\textrm{td}(d)}
    \end{bmatrix} 
    =
    \begin{bmatrix}
    \textbf{e}_\text{proc}[1] & \textbf{e}_\text{proc}[2] & \cdots & \textbf{e}_\text{proc}[N-d] \\
    \textbf{e}_\text{proc}[2] & \textbf{e}_\text{proc}[3] & \cdots & \textbf{e}_\text{proc}[N-d+1] \\
    \vdots & \vdots & \ddots & \vdots \\
    \textbf{e}_\text{proc}[d] & \textbf{e}_\text{proc}[d+1] & \cdots & \textbf{e}_\text{proc}[N-1] \\
    \textbf{e}_\text{proc}[d+1] & \textbf{e}_\text{proc}[d+2] & \cdots & \textbf{e}_\text{proc}[N]
    \end{bmatrix}.
    \)}
    \end{aligned}
    \label{eqn:hankel-lifting}
\end{equation}
The lifting in (\ref{eqn:hankel-lifting}) was applied to the downsampled EMG matrix \(E\) and the grip force matrix \(G\), using \(d=60\) time delays determined empirically. A complementary nonlinear lifting was used to map EMG plateaus and transitions to grip force patterns, producing gridded indicator observables. These observables, derived from time-delay embedded data, discretize the embedding space into a 3D grid to capture temporal dynamics. The grid is defined on a Cartesian plane with three time delays---\(\mathbf{e}_{\textrm{td}(1)}, \mathbf{e}_{\textrm{td}(1+\tau_1)},\, \mathbf{e}_{\textrm{td}(1+\tau_2)}\)---with each subregion treated as a binary observable (Fig.~\ref{fig:gridded-indicator-observables}). For any subregion \(S_{ijk}\), an indicator function assigns a value of one to points within the subregion and zero otherwise. Let \(S_{ijk}\) be a subregion in \([0, 1]^3\), defined by the lower (LO) and upper (UP) grid bounds \(\textbf{b}_{ijk} = \big[b_{i,\text{LO}}, b_{i,\text{UP}}, b_{j,\text{LO}}, b_{j,\text{UP}}, b_{k,\text{LO}}, b_{k,\text{UP}}\big]\):
\begin{itemize}
    \item For \(\mathbf{e}_{\textrm{td}(1)}\): lower limit \(b_{i,\text{LO}}\), upper limit \(b_{i,\text{UP}}\).
    \item For \(\mathbf{e}_{\textrm{td}(1+\tau_1)}\): lower limit \(b_{j,\text{LO}}\), upper limit \(b_{j,\text{UP}}\).
    \item For \(\mathbf{e}_{\textrm{td}(1+\tau_2)}\): lower limit \(b_{k,\text{LO}}\), upper limit \(b_{k,\text{UP}}\).
\end{itemize}
We conducted a targeted experimental investigation to optimize the grid for observables and minimize the model's estimation error. Since finer grids yield too many indicator observables that hinder real-time training, we limited the grid to 22 divisions \(\mathbf{b}_{ijk}\), yielding 21 subregions \(S_{ijk}\) per time delay. A power function (exponent \num{1.8}) was applied to adjust grid spacing, aligning processed EMG plateaus (red, Fig.~\ref{fig:proc-est-signal}) with \num{5} grip force levels (blue, Fig.~\ref{fig:proc-est-signal} and Table~\ref{tab:rbd-design}). The exponent was determined via regression analysis. The gridded indicator observable \(\mathbf{e}_{\mathrm{I},S_{ijk},\tau_1,\tau_2} \in \{0,1\}\) is defined as:
\begin{equation}
\begin{aligned}
    \mathbf{e}_{\mathrm{I},S_{ijk},\tau_1,\tau_2} = &
    \begin{cases}
      & \big(b_{i,\text{LO}} \leq \mathbf{e}_{\textrm{td}(1)} \leq b_{i,\text{UP}}\big) \, \text{ and }\\
      & \big(b_{j,\text{LO}} \leq \mathbf{e}_{\textrm{td}(1+\tau_1)} \leq b_{j,\text{UP}}\big) \,\text{ and } \\
      & \big(b_{k,\text{LO}} \leq \mathbf{e}_{\textrm{td}(1+\tau_2)} \leq b_{k,\text{UP}}\big),
    \end{cases} \\
    & \text{for} \ i, j, k = 0, \dots, 20.
\end{aligned}
\label{eqn:gridded-indicator-obs}
\end{equation}
Lifting in this manner can produce many empty (all-zero) or sparse (mostly zero) observables, potentially leading to overfitting of the estimation model. We applied a constraint to retain only observables with at least \qty{0.1}{\percent} density during algorithm testing to mitigate this. Gridding was performed simultaneously on time delays \(\mathbf{e}_{\textrm{td}(1)}\), \(\mathbf{e}_{\textrm{td}(30)}\), and \(\mathbf{e}_{\textrm{td}(60)}\) (\(\tau_1=29,\ \tau_2=59 \to \mathbf{e}_{\mathrm{I},S_{ijk},29,59}\)), initially producing \(21^3 = \num{9261}\) observables (Fig.~\ref{fig:gridded-indicator-observables}), most of which were discarded due to high sparsity. The final lifted data matrices were assembled by stacking time-delayed observables augmented with gridded indicators for processed EMG and zero rows for grip force.
\begin{equation}
    \resizebox{0.91\columnwidth}{!}{\(
    \begin{aligned}
    E & = \phi\big(\textbf{e}_\text{proc}\big) = \big( \mathbf{e}_{0}, \mathbf{e}_{\text{td}(1)}, \ldots, \mathbf{e}_{\text{td}(60)}, \ldots, \mathbf{e}_{\mathrm{I},S_{ijk},29,59}, \ldots \big)^T \\
    G & = \psi\big(\textbf{g}\big) = \big( \mathbf{g}_{0}, \mathbf{g}_{\text{td}(1)}, \ldots, \mathbf{g}_{\text{td}(60)}, \ldots, \mathbf{0}, \ldots \big)^T.
    \end{aligned}
    \)}
    \label{eqn:final-lifting-estimation}
\end{equation}
Using (\ref{eqn:static-koopman}) and (\ref{eqn:final-lifting-estimation}), the Koopman estimation model \(\overline{\mathcal{K}}_\text{e}\) was trained on the full \qtyrange{20}{30}{\s} calibration experiment in approximately \qty{1.5}{\s}. The resulting grip force approximations \(\textbf{g}_\text{e}\) were thresholded to a minimum value of \qty{-1}{\N}.

\subsubsection{Forecasting Grip Force}
\label{subsec:prediction}
We propose a novel methodology for short-term grip force forecasting over a future horizon of \qty{0.5}{\s}, leveraging previously estimated grip force \(\textbf{g}_\text{e}\) and the Koopman operator for dynamic systems in~\eqref{eq:dynamical_evolution}. The Koopman operator is well-suited for this task due to its proven adaptability in dynamically changing environments \cite{Mezic2021}, such as object grasping. Our goal is to determine the Koopman operator \(\mathcal{K}_\text{p}\) that predicts the system's evolution by mapping observables within the same space (\(\mathcal{K}_\text{p}: \mathcal{O}_\mathcal{N} \to \mathcal{O}_\mathcal{N}\)) for forecasting. This approach enables fast training on data mini-batches, ensuring adaptation to the system's latest state. To approximate the potentially infinite-dimensional \(\mathcal{K}_\text{p}\), we employ Dynamic Mode Decomposition (DMD)~\cite{Schmid2022} to compute its spectral properties as Ritz pairs (\(\lambda_j\) - Ritz values and \(\mathbf{z}_j\) - Ritz vectors).
\begin{equation}
\begin{aligned}
    \mathcal{K}_\text{p}Z & = Z\Lambda, \\
    Z & = \left( \mathbf{z}_1, \ldots, \mathbf{z}_j, \ldots, \mathbf{z}_r \right), \\
    \Lambda & = \diag{\left( \lambda_1, \ldots, \lambda_j, \ldots, \lambda_r \right)},
\end{aligned}
\label{eqn:dmd-ritz-pairs}
\end{equation}
where \(r\) represents the total number of Koopman modes after DMD. To solve \eqref{eqn:dmd-ritz-pairs} and obtain the spectral properties of the Koopman operator, we used a lifted input matrix of grip force estimations \(\textbf{g}_\text{e}\) and employed the pyKMD suite\footnote{\url{https://apps.aimdyn.com/}}, which implements the Refined Rayleigh-Ritz Data-Driven Modal Decomposition (DDMD\_RRR) method alongside QR compression \cite{Drmac2018}. Enhanced DDMD\_RRR refines Ritz vectors \(\mathbf{z}_j\) and improves spectral accuracy by minimizing residuals in a data-driven setting, while QR compression ensures efficient real-time execution. We applied Hankel-DMD time-delay embedding \cite{Arbabi2017} to \(\textbf{g}_\text{e}\), as done for the estimation Koopman operator in \eqref{eqn:hankel-lifting}, effectively capturing temporal dynamics for grip force prediction. The number of time delays was kept within \numrange{4}{10}, fine-tuned during optimization. Before lifting, \(\textbf{g}_\text{e}\) was smoothed using Locally Weighted Scatterplot Smoothing (LOWESS) \cite{Cleveland1979} with a single iteration and window size calculated as: (smoothing coefficient \(\times\) batch size) = (\numrange[range-phrase=\ to\ ]{1.1}{1.9} \(\times 496\)). This smoothing reduced spikes and prediction errors by incorporating data from current and previous batches, with the smoothing coefficient fine-tuned as a hyperparameter. 
Additional lifting produced an interaction (combination) matrix \(G_{\textrm{e,int}}\) from the \(\ln\)-transformed time-delay observables of \(\textbf{g}_\text{e}\) (see (\ref{eqn:hankel-lifting})) to reduce prediction error. A vertical shift of \(+\qty{10}{\N}\) was applied to ensure all values were positive for the \(\ln\) transformation. The interaction component of the forecast input matrix is:
\begin{equation}
    \resizebox{0.89\columnwidth}{!}{\(
    \begin{aligned}
        G_{\textrm{e,int}} = \begin{bmatrix}
        \ln{\textbf{g}_{\textrm{e}}[1]} \ln{\textbf{g}_{\textrm{e}}[2]} & \ln{\textbf{g}_{\textrm{e}}[2]} \ln{\textbf{g}_{\textrm{e}}[3]} & \cdots & \ln{\textbf{g}_{\textrm{e}}[N-d]} \ln{\textbf{g}_{\textrm{e}}[N-d+1]} \\
        \ln{\textbf{g}_{\textrm{e}}[1]} \ln{\textbf{g}_{\textrm{e}}[3]} & \ln{\textbf{g}_{\textrm{e}}[2]} \ln{\textbf{g}_{\textrm{e}}[4]} & \cdots & \ln{\textbf{g}_{\textrm{e}}[N-d]} \ln{\textbf{g}_{\textrm{e}}[N-d+2]} \\
        \vdots & \vdots & \ddots & \vdots \\
        \ln{\textbf{g}_{\textrm{e}}[d-1]} \ln{\textbf{g}_{\textrm{e}}[d+1]} & \ln{\textbf{g}_{\textrm{e}}[d]} \ln{\textbf{g}_{\textrm{e}}[d+2]} & \cdots & \ln{\textbf{g}_{\textrm{e}}[N-2]} \ln{\textbf{g}_{\textrm{e}}[N]} \\
        \ln{\textbf{g}_{\textrm{e}}[d]} \ln{\textbf{g}_{\textrm{e}}[d+1]} & \ln{\textbf{g}_{\textrm{e}}[d+1]} \ln{\textbf{g}_{\textrm{e}}[d+2]} & \cdots & \ln{\textbf{g}_{\textrm{e}}[N-1]} \ln{\textbf{g}_{\textrm{e}}[N]}
        \end{bmatrix}
    \end{aligned}
    \)}
    \label{eqn:log-interaction-lifting}
\end{equation}
The complete lifted input matrix is obtained by stacking \eqref{eqn:log-interaction-lifting} with the time-delay embedding of \(\mathbf{g}_\text{e}\), yielding a sequence of snapshots \(\mathbf{g}_{\textrm{e,lift}(i)}\):
\begin{equation}
    \begin{aligned}
    G_{\textrm{e,lift}} =
    \begin{pmatrix}
        G_{\textrm{e,int}} \\
        G_{\textrm{e,td}}
    \end{pmatrix} = \big( \mathbf{g}_{\textrm{e,lift}(1)}, \ldots, \mathbf{g}_{\textrm{e,lift}(N-d)} \big).
    \end{aligned}
    \label{eqn:lift-in-data}
\end{equation}
Before applying DMD to \eqref{eqn:lift-in-data}, a snapshot thinning step was performed by removing nearby columns in \(G_{\textrm{e,lift}}\) while preserving the time-delay structure, effectively reducing computational burden~\cite{Frame2023}. Thinning demonstrated no loss of accuracy during subsequent hyperparameter tuning within the range \numrange{3}{8}. Additionally, it enabled forecasting at \qtyrange{16}{41}{\Hz}, sufficient for capturing decision-making patterns during gripping~\cite{Miklashevsky2022}.
After obtaining the spectral decomposition of the Koopman operator \(\mathcal{K}_\text{p}\), we compute the Koopman amplitudes \(\boldsymbol{\alpha} = [\alpha_1, \alpha_2, \dots, \alpha_\ell]^\top \in \mathbb{C}^\ell\), to reconstruct the input data matrix or predict the next state. This step is formulated as a least-squares minimization problem:
\begin{equation}
    \begin{aligned}
    \underset{\boldsymbol{\alpha} \in \mathbb{C}^\ell}{\min} 
    \sum_{i=1}^{N-d} \left\| \mathbf{g}_{\textrm{e,lift}(i)} - \sum_{j=1}^{\ell} \mathbf{z}_j \alpha_j \lambda_j^{i-1} \right\|_2^2,
    \end{aligned}
    \label{eqn:coeffs-ls-min}
\end{equation}
where $\ell$ denotes the number of modes kept after mode reduction $r \to \ell$. Implemented solver from \cite{Drmac2020} can efficiently solve (\ref{eqn:coeffs-ls-min}) using normal equations or, if the problem is ill-conditioned, a QR factorization-based solver, both of which are contained within the pyKMD framework. After obtaining predictions, as a final step, too low or too high values were thresholded to the minimum and maximum grip force from the calibration experiment. Finally, forecasting future snapshots in horizon \(h\) can be approximated using:
\begin{equation}
    \begin{aligned}
    \mathbf{g}_{\textrm{e,lift}(N-d+h)} \approx \sum_{j=1}^{\ell} \mathbf{z}_j \alpha_j \lambda_j^{N-d+h-1}, \quad h = 1, \dots.
    \end{aligned}
    \label{eqn:predict-snapshot}
\end{equation}
The reported error metric for estimating and predicting grip force is the \textit{Weighted Mean Absolute Percentage Error} (wMAPE):
\begin{equation}
    \text{wMAPE} = \frac{\sum_{i=1}^{N} |\hat{g_i} - g_i|}{\sum_{i=1}^{N} |g_i|}
    \label{eqn:wmape}
\end{equation}
This relative metric was chosen because it effectively handles close-to-zero values by normalizing absolute errors with the sum of actual values, making it suitable for comparison across measurements with varying absolute grip magnitudes.

\begin{figure}[!t]
    \centering
    \includegraphics[width=\columnwidth]{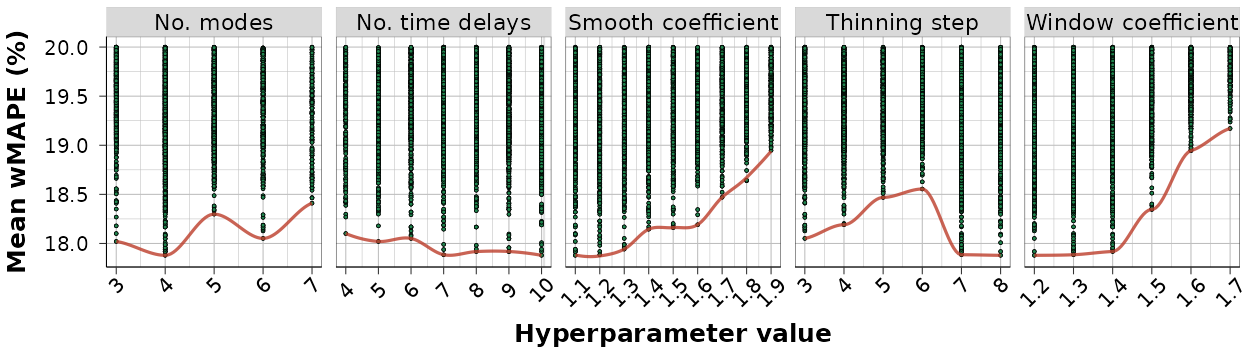}
    \caption{Hyperparameter tuning runs for optimizing prediction algorithm. wMAPE is an average of all predictions across all 52 experiments. A smooth red line connects the minimum run values for hyperparameter levels.}
    \label{fig:hyperparameter-tuning}
\end{figure}

Final hyperparameter tuning for the prediction model was conducted using a grid search across five parameters: number of Koopman modes after reduction \(\ell\), number of time delays \(d\), batch window modifier coefficient for smoothing, batch window modifier for prediction, and thinning step. Based on the hyperparameter tuning results shown in Fig.~\ref{fig:hyperparameter-tuning}, with runs featuring extremely high errors excluded, it can be concluded that the minimum wMAPE error is achieved with four Koopman modes, \numrange{7}{10} time delays, batch window modifier coefficient for smoothing in the range of \numrange{1.1}{1.2}, thinning step in the range of \numrange{7}{8}, and a batch window size modifier coefficient in the range of \numrange{1.2}{1.4}. To select the final hyperparameters for the prediction algorithm, the median wMAPE value across all measurements was also computed to enhance metric robustness. The hyperparameters resulting in the minimal sum of mean and median wMAPE are presented in Table~\ref{tab:hyperparameter-tuning}. The final error metric using these optimal hyperparameters is discussed in Section~\ref{sec:results}.
% Reduce spacing between columns in a table
\setlength{\tabcolsep}{3pt}
\begin{table}[!h]
\caption{Optimal hyperparameters for prediction model obtained after tuning results for prediction algorithm.}
\label{tab:hyperparameter-tuning}
\centering
\begin{tabulary}{\columnwidth}{CCCCC}
\toprule
\textbf{Batch window size modifier coefficient} & \textbf{Batch modifier coefficient for smoothing} & \textbf{Thinning step}  & \textbf{No. time delays} & \textbf{No. Koopman modes} \\
\midrule
1.3 & 1.1 & 7 & 8 & 4 \\ 
\bottomrule
\end{tabulary}
\end{table}

\section{Results \& Experimental Validation}
\label{sec:results}
\begin{figure}[!t]
    \centering
    \includegraphics[width=\columnwidth]{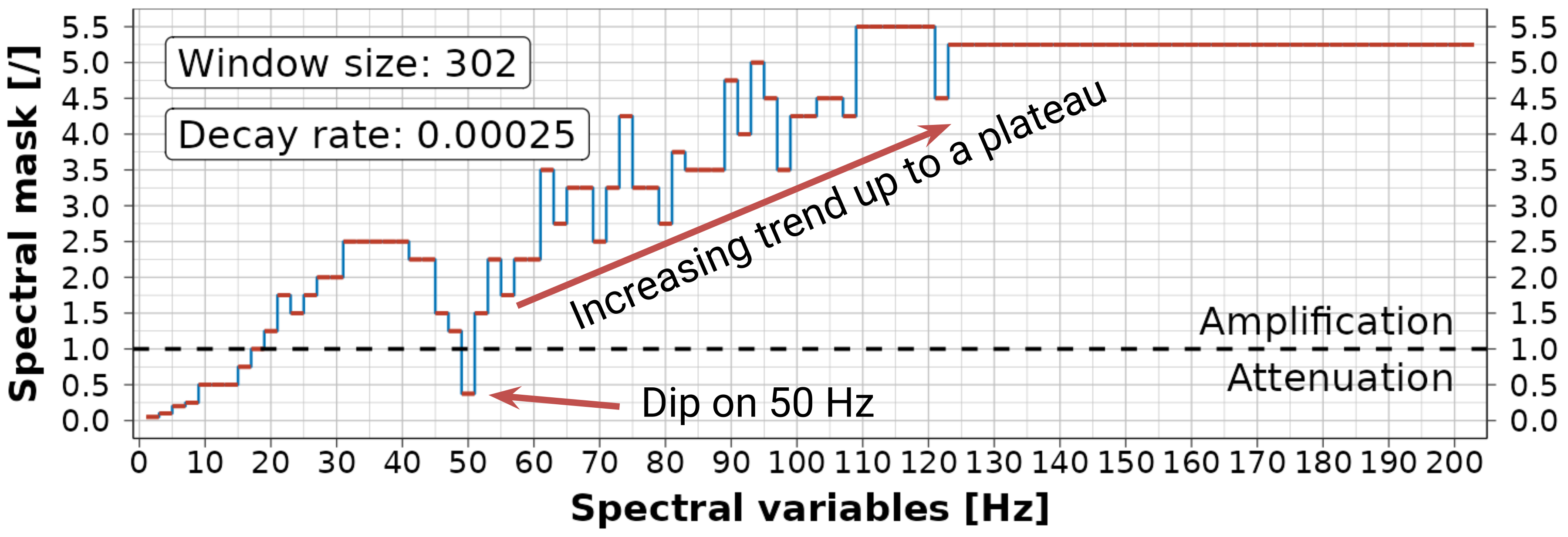}
    \caption{Optimal spectral mask for processing EMG signal.}
    \label{fig:spectral-mask}
\end{figure}

The analysis of the most sensitive frequencies from the multi-step SA (Fig.~\ref{fig:frequency-bounds-steps}) shows the need to first act on reducing the lower frequency spectral components ($\leq$\qty{14}{Hz}) and those in the \qtyrange{46}{50}{\Hz} range to improve the peak cross-correlations. Once the bounds of these initial sensitive spectral components are narrowed to near-optimal values, the range of sensitive frequencies expands, filling the gap between \qtyrange{14}{46}{\Hz} and extending towards higher-frequency components (arrows in Fig.~\ref{fig:frequency-bounds-steps}).
Typically, parameter optimization would be performed within the narrowed decision vector bounds. However, it is unnecessary in this case due to the descriptive statistics of the resulting mean peak cross-correlation computed from all LH samples after the final SA step. Descriptive statistics for different positions are:
\begin{itemize}
    \item P1: mean \num{0.956}, SD \num{4.13e-4}, min. \num{0.954}, max. \num{0.958},
    \item P2: mean \num{0.960}, SD \num{5.35e-4}, min. \num{0.958}, max. \num{0.962}.
\end{itemize}
Because the standard deviation of the mean peak cross-correlations indicates extremely low variability, the average values between each decision variable’s upper and lower bounds will be used as the optimal mask values, as shown in Fig.~\ref{fig:spectral-mask}. Sections of the optimal spectral mask align with findings from \cite{Konrad2006}, where lower-frequency components associated with wire movements are either attenuated or removed. The attenuation is nearly linear, with DC component and \qty{2}{Hz} fully filtered out, \qty{10}{Hz} reduced by \qty{50}{\percent}, and \qty{18}{Hz} left unaffected. Frequencies from \qtyrange{20}{48}{\Hz} require amplification following an inverted U-shaped curve, starting and ending at \qty{25}{\percent}, with a peak amplification of \qty{150}{\percent} between \qtyrange{32}{42}{\Hz}.

The inverted U is followed by a sharp dip at the \qty{50}{\Hz} component, indicating that electrical ground noise impacts the recorded signal. The \qty{50}{\Hz} spectral component proves particularly challenging to process, as neither retaining it at nominal amplitudes (with a BP filter) nor completely removing it (using a notch filter) is sufficient. Our optimizations show that maintaining \qty{50}{\Hz} at \qty{37.5}{\percent} of its amplitude preserves enough of the signal’s power for grip force modeling. This finding supports the conclusions of \cite{Konrad2006}, which state that a significant portion of the EMG signal's power spectrum is concentrated at \qty{50}{\Hz} and should not be entirely filtered out.

The next section of the mask targets amplifying the mid-frequency spectra, where most of the signal's power is concentrated. As the amplitudes of the spectral components decrease, the mask progressively amplifies them, increasing approximately linearly from \qty{50}{\%} at \qty{52}{\Hz} to \qty{450}{\%} at \qty{110}{\Hz}. This amplification trend plateaus at \qtyrange{425}{450}{\percent} around \qty{110}{\Hz} and extends up to \qty{202}{\Hz}. A possible explanation for the need to amplify mid-frequency spectral components may lie in three physiological phenomena related to the spatial low-pass filtering of recorded EMG signals: MU structure, volume conduction, and electrode positioning \cite{Stegeman2000}. First, the MU structure introduces spatial smoothing at the signal source due to the scattered arrangement of muscle fibers within a MU. Volume conduction refers to the signal's attenuation and distortion as it propagates through biological tissues from the source to the skin surface, where it is measured. Finally, electrode positioning contributes to low-pass filtering, as the electrodes capture an averaged signal over the area they cover, further reducing higher-frequency content.

We conducted a stepwise ablation study on the spectral mask plateau (\qtyrange{110}{202}{\Hz}) to validate our conclusions. We systematically excluded frequency bands in \qty{4}{\Hz} increments, starting from \qtyrange{110}{202}{\Hz} and progressing to \qtyrange{198}{202}{\Hz}. Each step resulted in a lower-than-optimal mean peak cross-correlation between processed EMG and measured grip force. 

Our SA revealed the minimal influence of higher-frequency components (\qtyrange{204}{498}{\Hz}). We performed an additional ablation study by varying the mask from \numrange{0}{5} in this range. As no significant changes in mean peak cross-correlation were observed, we set the mask to 0 for the \qtyrange{204}{498}{\Hz} range, effectively shielding the system from potential noise in this band. This approach to processing sEMG signals is innovative, and some of the resulting observations warrant further investigation for a clearer understanding.

\begin{figure*}[!ht]
    \centering
    \begin{subfigure}{\textwidth}
        \includegraphics[width=0.95\textwidth]{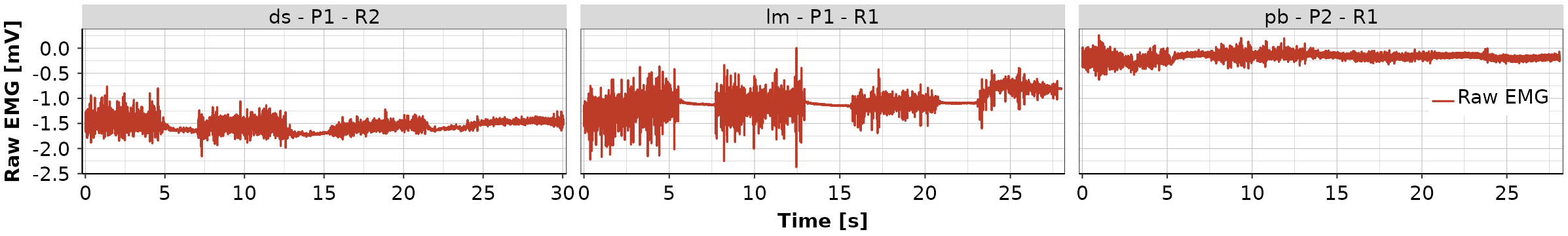}
        \caption{Examples - measured raw EMG signal.}
        \label{fig:raw-emg-signal}
    \end{subfigure}
    \begin{subfigure}{\textwidth}
        \includegraphics[width=\textwidth]{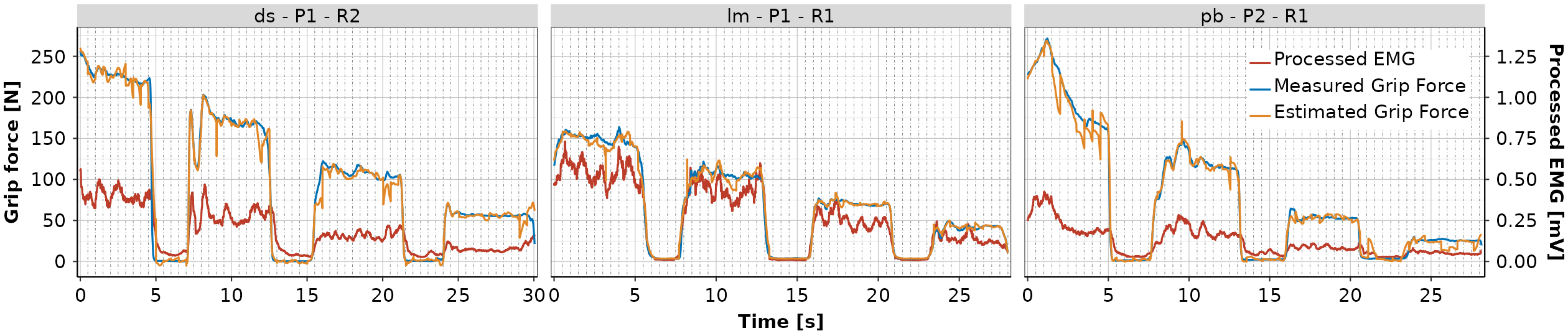}
        \caption{Examples - processed EMG signal and comparison of estimated and measured grip force.}
        \label{fig:proc-est-signal}
    \end{subfigure}
    \begin{subfigure}{\textwidth}
        \includegraphics[width=0.95\textwidth]{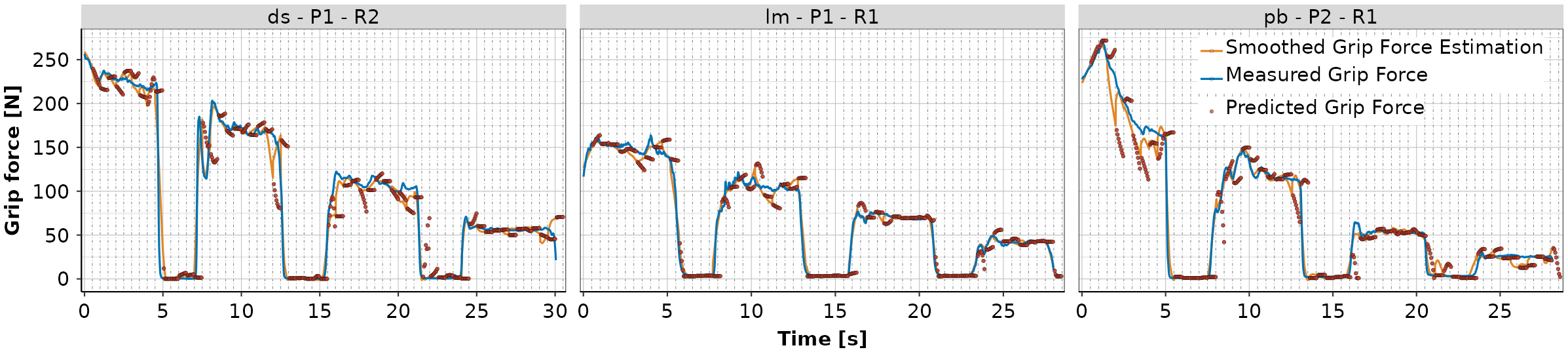}
        \caption{Examples - predicted grip force from smoothed estimation.}
        \label{fig:pred-smooth-signal}
    \end{subfigure}
    \caption{Examples of all signals relevant for grip force modeling - raw and processed EMG signals, estimates, smoothed estimates, forecasts, and measured grip force.}
    \label{fig:emg-grip-signals}
\end{figure*}

By applying the signal processing methods outlined in Section~\ref{sec:sensitivity_analysis}, the processed EMG signals, shown in red in Fig.~\ref{fig:proc-est-signal}, demonstrate a clear high cross-correlation with the measured grip force. Table~\ref{tab:correlations-summary} summarizes peak cross-correlations for all measurements at both sensing positions, showing very strong correlations. The analysis reveals that EMG signals lag behind grip force measurements, with delays ranging from \qtyrange{0}{156}{\ms}.

% Reduce spacing between columns in a table
\setlength{\tabcolsep}{2.5pt}
\begin{table}[!t]
\caption{Cross-correlation summary statistics for all measurements on P1 and P2: Optimal processed EMG -- grip force.}
\label{tab:correlations-summary}
\centering
\begin{tabulary}{\columnwidth}{RRRRRRR}
  \toprule
& \bfseries{Min.} & \bfseries{1st Qu.} & \bfseries{Median} & \bfseries{Mean} & \bfseries{3rd Qu.} & \bfseries{Max.} \\
  \midrule
\multicolumn{7}{l}{\bfseries{Position 1}} \\
Peak cross-correlation & 0.891 & 0.947 & 0.964 & 0.958 & 0.971 & 0.987 \\
Optimal time lag [ms] & 0.0 & 0.0 & 43.8 & 54.0 & 96.4 & 156.1 \\
   \midrule
\multicolumn{7}{l}{\bfseries{Position 2}} \\
Peak cross-correlation & 0.924 & 0.952 & 0.967 & 0.962 & 0.972 & 0.988 \\
Optimal time lag [ms] & 0.0 & 0.0 & 16.1 & 25.0 & 42.8 & 78.5 \\
   \bottomrule
\end{tabulary}
\end{table}

The RBD experimental design allows for investigating the effects of subject and sensing location on the wMAPE estimation metric. Using the procedure described in Section~\ref{subsec:estimation}, wMAPE was calculated for 52 experiment runs (see Table~\ref{tab:rbd-estimate-wmape}). The mean estimation wMAPE of \qty{5.5}{\%} reflects fit error, as the model was trained on each subject’s full dataset and evaluated on \qty{0.5}{s} batches without a test set, given Koopman operators’ deterministic linearity. These results demonstrate effective grip force estimation from a single sensing position. In addition to single-measurement error metrics, we computed means and effects across blocks containing all measurements for each position and subject. The effects were computed as the difference between the block mean and the overall mean wMAPE. Although the means and effects varied considerably across subjects, the mean for P1 was only \qty{0.2}{\%} worse than the overall mean, while the mean for P2 was \qty{0.2}{\%} better.
% Reduce spacing between columns in a table
\setlength{\tabcolsep}{2.2pt}
% After autogenerate: R11 -> R1, R21 -> R2, delete four zeros, paste midrule before Subject, reduce multicolumn to 14, and add l||
% latex table generated in R 4.4.0 by xtable 1.8-4 package
% Tue Aug 13 05:00:36 2024
\begin{table}[!h]
\caption{wMAPE (\%) for estimating grip force from EMG per subject and position. (R: Replication; M: Mean; E: Effect)}
\label{tab:rbd-estimate-wmape}
\centering
\begingroup\scriptsize
\begin{tabulary}{\columnwidth}{LRRRRRRRRRRRRRRR}
  \toprule
& \multicolumn{13}{c}{\bfseries{\small Subject}} & \multicolumn{2}{c}{\bfseries{\small Position}} \\
\cmidrule(lr){2-14}
\cmidrule(lr){15-16}
 & \small ac & \small dp & \small ds & \small js & \small lb & \small lk & \small lm & \small ln & \small md & \small mm & \small nk & \small pb & \small ss & \small M & \small E \\ 
  \midrule
\multicolumn{14}{l}{\bfseries{\small Position 1}}\\
\footnotesize R1 & 4.4 & 6.4 & 10.0 & 4.5 & 7.7 & 6.6 & 4.5 & 8.4 & 4.3 & 5.1 & 2.7 & 6.1 & 3.8 & & \\ 
  \footnotesize R2 & 2.3 & 4.7 & 5.4 & 3.9 & 8.0 & 6.9 & 7.6 & 3.7 & 5.2 & 6.8 & 6.4 & 7.9 & 4.3 & 5.7 & 0.2 \\ 
   \midrule
\multicolumn{14}{l}{\bfseries{\small Position 2}}\\
\footnotesize R1 & 4.0 & 4.1 & 4.7 & 3.9 & 3.3 & 5.8 & 3.6 & 4.4 & 7.2 & 8.5 & 3.3 & 6.7 & 4.4 & & \\ 
  \footnotesize R2 & 3.8 & 4.3 & 4.4 & 5.1 & 4.9 & 6.2 & 4.4 & 4.6 & 9.2 & 9.9 & 3.2 & 10.1 & 3.3 & 5.3 & -0.2 \\ 
   \midrule
   \midrule
\multicolumn{14}{l}{\bfseries{\small Subject mean and effect}}\\
\footnotesize M & 3.6 & 4.9 & 6.1 & 4.4 & 6.0 & 6.4 & 5.1 & 5.3 & 6.5 & 7.6 & 3.9 & 7.7 & 4.0 & & \\ 
  \footnotesize E & -1.8 & -0.6 & 0.6 & -1.1 & 0.5 & 0.9 & -0.4 & -0.2 & 1.0 & 2.1 & -1.6 & 2.2 & -1.5 & & \\ 
   \bottomrule
\multicolumn{16}{l}{\rule{0em}{2.5ex}\small Overall estimation mean wMAPE: 5.48$\%$}\\
\end{tabulary}
\endgroup
\end{table}

An analysis of variance (ANOVA) was conducted on the blocked RBD with estimation wMAPE to assess the significance of the position effect using a \qty{5}{\percent} significance level. The results, presented in Table~\ref{tab:rbd-aov}, show that, while the effect of the subject on wMAPE was significant (\(p\)-value = \num{0.015}), the effect of position (when the subject effect is removed) was not (\(p\)-value = \num{0.422}). It can be concluded that the placement of EMG electrodes along the flexor carpi ulnaris muscle, whether in position \num{1} or \num{2}, does not significantly influence the estimation error.
Three representative examples of estimated grip force, with errors corresponding approximately to the first, second, and third quartiles, are highlighted in yellow in Fig.~\ref{fig:proc-est-signal}. The complete set of graphs is available in the first author's GitHub repository\footnote{\url{https://github.com/tbazina/grip-emg-optimize}}.
% Reduce spacing between columns in a table
\setlength{\tabcolsep}{1.4pt}
% latex table generated in R 4.4.0 by xtable 1.8-4 package
% Wed Aug 14 11:22:02 2024
\begin{table}[!t]
\caption{RBD results - non-significant effect of measuring position on wMAPE.} 
\label{tab:rbd-aov}
\centering
\begingroup\scriptsize
\begin{tabulary}{\columnwidth}{LRRRRRRRRRR}
  \toprule
& \multicolumn{5}{c}{\bfseries{\small Estimation}} & \multicolumn{5}{c}{\bfseries{\small Prediction}} \\
\cmidrule(lr){2-6}
\cmidrule(lr){7-11}
 & Df & Sum Sq & Mean Sq & F value & Pr($>$F) & Df & Sum Sq & Mean Sq & F value & Pr($>$F) \\
  \midrule
Position    & 1 & 1.91 & 1.91 & 0.66 & 0.422 & 1 & 0.35 & 0.35 & 0.03 & 0.853 \\ 
  Subject     & 12 & 87.51 & 7.29 & 2.52 & 0.015 & 12 & 232.00 & 19.33 & 1.94 & 0.060 \\ 
  Resids   & 38 & 110.15 & 2.90 &  &  & 38 & 378.55 & 9.96 &  & \\ 
   \bottomrule
\end{tabulary}
\endgroup
\end{table}

Following a similar approach to the estimations, the methodology from Section~\ref{subsec:prediction} was applied to short-term grip force forecasting in \qty{0.5}{s} mini-batches over a \qty{0.5}{s} horizon. The adaptive forecasting model was trained on the previous $1.3 \times \qty{0.5}{s}$ batches and tested on the subsequent $1 \times \qty{0.5}{s}$ batch. wMAPE values were computed for all measurements and reported as RBD results in Table~\ref{tab:rbd-predict-wmape}. Using the optimal hyperparameters from Table~\ref{tab:hyperparameter-tuning}, we achieved approximately \qty{17.9}{\percent} overall forecast wMAPE. While the effect of the subject varied, the effect of position on wMAPE remained within the \qty{\pm 0.1}{\percent} range. The ANOVA analysis in Table~\ref{tab:rbd-aov} confirmed the non-significant effect of position ($p = \num{0.853}$) and subject ($p = \num{0.06}$) on forecast error, further demonstrating robustness to electrode placement along the flexor carpi ulnaris muscle. Examples of forecasts with error metric values approximately corresponding to the first, second, and third quartiles are shown as red dots in Fig.~\ref{fig:pred-smooth-signal}, with the smoothed grip force estimation, which serves as the input signal for forecasting, displayed in yellow.
% Reduce spacing between columns in a table
\setlength{\tabcolsep}{1.6pt}
% latex table generated in R 4.4.1 by xtable 1.8-4 package
% Fri Sep 13 14:25:01 2024
\begin{table}[!h]
\caption{wMAPE (\%) for predicting grip force from EMG per subject and position. (R: Repetition; M: Mean; E: Effect}
\label{tab:rbd-predict-wmape}
\centering
\begingroup\scriptsize
\begin{tabulary}{\columnwidth}{LRRRRRRRRRRRRRRR}
  \toprule
& \multicolumn{13}{c}{\bfseries{\small Subject}} & \multicolumn{2}{c}{\bfseries{\small Position}} \\
\cmidrule(lr){2-14}
\cmidrule(lr){15-16}
 & \small ac & \small dp & \small ds & \small js & \small lb & \small lk & \small lm & \small ln & \small md & \small mm & \small nk & \small pb & \small ss & \small M & \small E \\ 
  \midrule
\multicolumn{14}{l}{\bfseries{\small Position 1}}\\
\footnotesize R1 & 24.7 & 15.0 & 25.0 & 18.7 & 21.4 & 21.1 & 13.1 & 22.7 & 12.5 & 15.4 & 19.4 & 18.8 & 13.4 & & \\ 
  \footnotesize R2 & 23.6 & 15.4 & 17.9 & 17.6 & 15.0 & 18.9 & 17.5 & 15.1 & 22.3 & 14.7 & 18.6 & 19.6 & 10.6 & 18.0 & 0.1 \\ 
   \midrule
\multicolumn{14}{l}{\bfseries{\small Position 2}}\\
\footnotesize R1 & 14.8 & 17.2 & 16.7 & 17.7 & 16.1 & 20.4 & 12.9 & 18.5 & 15.8 & 17.2 & 19.1 & 20.2 & 15.2 & & \\ 
  \footnotesize R2 & 16.8 & 16.1 & 17.6 & 22.3 & 14.2 & 22.9 & 12.3 & 17.2 & 20.9 & 25.2 & 19.0 & 20.5 & 16.7 & 17.8 & -0.1 \\ 
   \midrule
   \midrule
\multicolumn{14}{l}{\bfseries{\small Subject mean and effect}}\\
\footnotesize M & 20.0 & 15.9 & 19.3 & 19.1 & 16.7 & 20.9 & 13.9 & 18.4 & 17.9 & 18.2 & 19.0 & 19.8 & 14.0 & & \\ 
  \footnotesize E & 2.1 & -2.0 & 1.4 & 1.2 & -1.3 & 2.9 & -4.0 & 0.4 & -0.0 & 0.2 & 1.1 & 1.8 & -4.0 & & \\ 
   \bottomrule
\multicolumn{16}{l}{\rule{0em}{2.5ex}\small Overall prediction mean wMAPE: 17.92$\%$}\\
\end{tabulary}
\endgroup
\end{table}

Fig.~\ref{fig:emg-grip-signals} contrasts all described sEMG and grip signals, from raw and processed EMG to estimated and measured grip force, resulting in smoothed estimates and short-term batch predictions.

\section{Conclusion}
\label{sec:conclusion}

This work focuses on developing real-time procedures for sensing hand muscle activity and estimating grip force, with potential applications in controlling rehabilitation devices. The methodology involves sensing muscle activity (physiological signals) using non-invasive surface electromyography (sEMG) sensors. The signals are cross-correlated with synchronously collected hand grip force, determined using a force-calibrated hand dynamometer as part of a properly designed experiment. We devised the composition of the sEMG signal processing steps to achieve a strong mean peak cross-correlation (\qty{\sim 96}{\percent}) between the EMG signal and grip force. We optimized decision variables (spectral mask, smoothing window size, and decay rate) to maximize cross-correlation. Finally, using processed EMG from a single sensing location, the static Koopman operator enables grip force estimation, while the dynamic operator facilitates short-term prediction. The algorithm follows these steps: processing raw EMG, transforming and lifting the data for estimation, applying the estimation model, smoothing the estimates and lifting them for prediction, continuously retraining the prediction model online, and generating forecasts.

While creating a single large model that would generalize across different individuals and sessions would be challenging due to the inherent variability in EMG signals \cite{Stegeman2000}, the proposed Koopman methodology requires less than \qty{30}{s} to perform a one-time, patient- and session-specific calibration experiment using both sEMG sensors and a hand dynamometer, followed by \qty{1.5}{s} for estimation model training. Experimental assessment demonstrated the forecasting model’s rapid training and execution, requiring only \qty{30}{ms} to generate \qty{0.5}{s} predictions after receiving the latest \qty{0.5}{s} data mini-batch. This confirms the framework’s readiness for real-time implementation. Our method demonstrated robustness across sensing locations, wMAPE of \qty{5.5}{\percent} for estimations and \qty{17.9}{\percent} for predictions. To contextualize, we compared our results to a baseline \cite{Li2014} which uses frequency-domain BP filtering (STFT with \qty{0.5}{s} window and \qty{50}{\percent} overlap), RMS smoothing (\qty{0.5}{s} window), and ordinary least squares regression. This baseline showed significantly higher wMAPE of \qty{24.4}{\percent} for estimations and requires an extra FFT-to-IFFT step, complicating real-time implementation.

Future work will integrate developed Robot Operating System (ROS) modules for sensor interaction, data acquisition, and signal processing with the Koopman real-time force estimation and prediction module. Advancements in simplifying hand kinematics from \cite{Bazina2024} will be incorporated to develop a comprehensive system, including a human agent model with various grasp modalities. This approach aims to accelerate the development of robotic rehabilitation devices, enhancing their effectiveness and adaptability to individual patient needs.

\appendix
\section{Appendix}
\label{app:appendix}
\subsection{Experiment Setup}
\label{app:experiment_setup}
\subsubsection{Hardware Components}
\label{app:hardware}
Shimmer3 EMG Unit\footnote{\scriptsize\url{https://shimmersensing.com/product/shimmer3-emg-unit/}}  was used for muscle activity sensing. A maximum internal gain of 12 was set, and an internal calibration procedure was followed. Two electrodes at a single sensing location, along with a third reference electrode, were used to address the challenge of the small EMG signal relative to noise. This setup employs Common Mode Rejection to eliminate shared noise through signal subtraction, effectively preserving and amplifying the localized EMG signal. For grip force measuring, a Vernier Go Direct® Hand Dynamometer\footnote{\scriptsize\url{https://vernier.com/product/go-direct-hand-dynamometer/}} was used, with a force range \qtyrange{0}{550}{\N}, resolution \qty{0.05}{\N}, and uncertainty \qty{1.96}{SD}.

\subsubsection{Dynamometer Calibration}
\label{app:calibration}
Dynamometer calibration was performed in-house according to the ASTM E74 norm. Calibration weights of OIML classes F1, M1, and M3 were used, and the laboratory temperature was maintained at \qty{\sim 23}{\degreeCelsius}. Calibration forces spanned: \qtylist{5;20;50;100;150;200;250;300;350;400;450;500;550}{\N}. The instrument was first preloaded from min to max force to establish hysteresis. Each measurement was preceded by reducing the instrument to zero between successive loadings, and the entire process was repeated three times. For measured forces greater than \qty{50}{\N}, the uncertainty as a percentage of the measured force is less than \qty{4}{\percent}, while for forces below \qty{50}{\N}, the percentage can be as high as \qty{10}{\percent}. The obtained calibration equation that relates the raw dynamometer signal \(g_\textrm{r}\) to the reported grip force \(g\) is:

\vspace{-0.5em}
\begingroup\footnotesize
\begin{equation}
    g = 1.063 g_\textrm{r} - 2.588 \times 10^{-4} g_\textrm{r}^2 - 9.003 \times 10^{-8} g_\textrm{r}^3 
    + 7.615 \times 10^{-10} g_\textrm{r}^4
    \label{eqn:calibration}
\end{equation}
\endgroup
The most prominent term in (\ref{eqn:calibration}) is linear, and \num{\sim 1}, due to demands of the ASTM E74 norm, a fourth-degree polynomial was deemed necessary. The ROS and Python were utilized to integrate devices and enable the acquisition of time-series raw EMG and calibrated dynamometer signals. The implemented modules are available in the first author's GitHub repositories, namely \texttt{godirect\_ros}\footnote{\url{https://github.com/tbazina/godirect_ros}} and \texttt{shimmer\_ros}\footnote{\url{https://github.com/tbazina/shimmer_ros}}.

\subsection{Sensitivity Analysis and Parameter Optimization}
\label{app:Sensitivity}
\subsubsection{Peak Cross-correlation}
\label{app:peak_cross_correlate_emg_grip}
Algorithm for computing peak cross-correlation between processed EMG signal \(\textbf{e}_\textrm{proc}\) and grip force \(\textbf{g}\) from single experiment run:
\begin{algorithm}[!h]
\caption{\( \PeakCrossCorr \big( \textbf{e}_\textrm{proc}, \textbf{g} \big) \)}
\label{alg:cross_correlate_emg_grip}
\algsetup{indent=0.8em, linenosize=\scriptsize}
\begin{algorithmic}
\REQUIRE \(\len{\left(\textbf{e}_\textrm{proc}\right)} == \len{\left(\textbf{g}\right)} \) \COMMENT{Equal in length}
\STATE \(T \leftarrow \len{\left(\textbf{e}_\textrm{proc}\right)}\) \COMMENT{Assign signal length}
\STATE \( \textbf{e}_{\text{proc,cent}} \leftarrow \textbf{e}_\textrm{proc} - \mean{\left(\textbf{e}_\textrm{proc}\right)} \) \COMMENT{Mean center EMG}
\STATE \( \textbf{g}_{\text{cent}} \leftarrow \textbf{g} - \mean{\left(\textbf{g}\right)} \) \COMMENT{Mean center grip}
\STATE \( \textbf{r}[k] \leftarrow \sum_{t=0}^{T-k-1} \textbf{e}_{\text{proc,cent}}[t] \cdot \textbf{g}_{\text{cent}}[t+k], \enspace k = 0, \dots, T-1 \) \COMMENT{Cross-correlation}
\STATE \( \textbf{r}_\text{norm} \leftarrow \frac{ \textbf{r} }{ T \cdot \SD{\left(\textbf{e}_{\text{proc,cent}}\right)} \cdot \SD{\left(\textbf{g}_{\text{cent}}\right)}} \) \COMMENT{Normalize cross-correlation to \( \left[ -1,1 \right] \)}
\STATE \( k_{\max} \leftarrow \underset{k}{\arg\max} \big(\abs{\left(\textbf{r}_\text{norm}\right)}\big) \) \COMMENT{Time lag at peak}
\RETURN \( \textbf{r}_{\text{norm}}[k_{\max}] \) \COMMENT{Peak cross-correlation}
\end{algorithmic}
\end{algorithm}

\subsubsection{Batch Processing EMG Signal}
\label{app:batch_process_emg}
Algorithm for computing processed EMG signal \(\mathbf{e}_{\text{proc}}\) from raw EMG data \(\mathbf{e}_\text{raw}\) by sequentially processing mini-batches \(\mathbf{e}_\textrm{raw,batch}\) (496 points each) using optimized spectral mask \(\mathbf{w}_\textrm{mask}\), exponential moving average window size \(w_\text{EMA}\), and decay factor \(\alpha_\text{EMA}\).
\begin{algorithm}[!t]
\caption{\hspace{-0.1em}\( \BatchProcessEMG \big( \textbf{e}_\textrm{raw}, \textbf{w}_\textrm{mask}, w_\text{EMA}, \alpha_\text{EMA} \big)\)}
\label{alg:processing_emg}
\algsetup{indent=0.8em, linenosize=\scriptsize}
\begin{algorithmic}
\STATE \(\textbf{e}_\text{proc} \leftarrow \left(\,\right)\) \COMMENT{Initialize empty tuple}
\FOR{each mini-batch \(\textbf{e}_\textrm{raw,batch}\) in entire \(\textbf{e}_{\textrm{raw}}\)}
\STATE \(T \leftarrow \len{\left(\textbf{e}_\textrm{raw,batch}\right)}\) \COMMENT{Assign batch length: \num{496}}
\STATE \( \textbf{e}_\textrm{FFT,batch} \leftarrow \mathcal{F} \big( \textbf{e}_\textrm{raw,batch} \big) \) \COMMENT{FFT of raw EMG signal}
\STATE \( \textbf{e}_{\text{mask,batch}} \leftarrow \mathcal{F}^{-1}\big( \textbf{e}_\textrm{FFT,batch} \odot \textbf{w}_\textrm{mask} \big) \) \COMMENT{Masking + iFFT}
\STATE \( \textbf{e}_{\text{abs,batch}} \leftarrow \abs{\left(\textbf{e}_{\text{mask,batch}}\right)} \) \COMMENT{Rectify signal}
\STATE \( \textbf{wnd}_{\text{EMA}}[i] \leftarrow \frac{(1 - \alpha_\text{EMA})^{i}}{\sum_{j=0}^{w_\text{EMA}-1} (1 - \alpha_\text{EMA})^j}, \, i = 0, \dots, w_\text{EMA}-1 \) \COMMENT{Create and normalize exponentially decaying window}
\STATE \( \textbf{e}_{\text{EMA,batch}}[t] \leftarrow \sum_{i=0}^{w_\textrm{EMA}-1} \textbf{wnd}_{\text{EMA}}[i] \odot \textbf{e}_{\text{abs,batch}}[t - i], \enspace t = 0, \dots, T-1\) \COMMENT{Convolve rectified signal with window. Use data from previous batch for negative \(t-i\).}
\STATE \(\textbf{e}_{\text{proc}} \leftarrow \left(\textbf{e}_{\text{proc}}, \textbf{e}_{\text{EMA,batch}}\right)\) \COMMENT{Concatenate batches}
\ENDFOR 
\RETURN \(\textbf{e}_{\text{proc}}\) \COMMENT{Return processed EMG signal}
\end{algorithmic}
\end{algorithm}

\subsubsection{Preliminary SA}
\label{app:preliminary_sa}
Grouped Sobol variance-based SA was conducted independently for P1 and P2 EMG measurements. For each position, $2^{16}$ samples of 250-dimensional decision vectors were generated using Saltelli's sampling scheme. Bootstrapping with $2^{16}$ resamples was applied to account for uncertainty, creating additional datasets by sampling the original one with replacement. SIs were calculated for each bootstrapped dataset, and empirical distributions were generated. From these distributions, \qty{95}{\percent} confidence intervals were derived.

\subsubsection{Iterative Multi-step Ungrouped SA}
\label{app:multistep_sa} RBD-FAST SA method returned only first-order SIs, with $2^{16}$ decision vector samples and resampling with bootstrapping using \num{8192} (\(2^{13}\)) samples for CI computation. After each step, the projections on the most sensitive spectral components, or smoothing parameters, were visualized using LH sampling with \num{10000} samples, and manual bounds narrowing was performed. The first step identified the \qty{2}{\Hz} spectral component as the most influential on cross-correlation variance, with an SI in the \qtyrange{78}{83}{\percent} range. The LH sample projection onto the 2 Hz component showed that increasing its mask variable significantly reduces the mean peak cross-correlation, prompting the narrowing of its variable bounds to the \numrange{0}{0.5} range. Additionally, LH sampling was projected onto the next three spectral components with the highest SI from the first SA step. Similar trends to those observed for the 2 Hz component were noted, prompting the narrowing of bounds to \numrange{0}{1} for \qty{4}{\Hz}, \numrange{0}{2} for \qty{6}{\Hz}, and \numrange{0}{3} for \qty{50}{\Hz}. Another \num{20} steps (\(2^\text{nd}-21^\text{th}\)) of iterative SA and optimization were performed similarly (see Fig.~\ref{fig:frequency-bounds-steps}). Figures displaying LH sampling projections, along with smooth and mean lines that support the reasoning behind the narrowing of bounds, are in the first author's GitHub repository\footnote{\url{https://github.com/tbazina/grip-emg-optimize}}.

\section*{Acknowledgment}
The authors thank students for conducting initial experiments and providing the experimental data. This work is supported by the Air Force Office of Scientific Research under award number FA9550-22-1-0531 and by the University of Rijeka under Grant uniri-iskusni-tehnic-23-47.

\printbibliography

@Article{Tarantola2006,
  author       = {Tarantola, S. and Gatelli, D. and Mara, T.A.},
  date         = {2006-06},
  journaltitle = {Reliab. Eng. Syst. Saf.},
  title        = {Random balance designs for the estimation of first order global sensitivity indices},
  issn         = {0951-8320},
  number       = {6},
  pages        = {717--727},
  volume       = {91},
  publisher    = {Elsevier BV},
}

@Article{Cutkosky1989,
  author       = {M.R. Cutkosky},
  date         = {1989-06},
  journaltitle = {IEEE Trans. Robot. Automat.},
  title        = {On grasp choice, grasp models, and the design of hands for manufacturing tasks},
  number       = {3},
  pages        = {269--279},
  volume       = {5},
  publisher    = {Institute of Electrical and Electronics Engineers ({IEEE})},
}

@Article{Feix2016,
  author       = {Feix, Thomas and Romero, Javier and Schmiedmayer, Heinz-Bodo and Dollar, Aaron M. and Kragic, Danica},
  year         = {2016},
  journal      = {IEEE Trans. Hum.-Mach. Syst.},
  title        = {The GRASP Taxonomy of Human Grasp Types},
  issn         = {2168-2305},
  number       = {1},
  pages        = {66--77},
  volume       = {46},
  publisher    = {Institute of Electrical and Electronics Engineers (IEEE)},
}

@Book{Konrad2006,
  author    = {Konrad, Peter},
  title     = {The ABC of EMG - A practical introduction to kinesiological electromyography},
  isbn      = {0-9771622-1-4},
  number    = {2006},
  pages     = {30--5},
  publisher = {Noraxon U.S.A. Inc.},
  volume    = {1},
  year      = {2006},
}

@article{Mezic2021,
  title={Koopman operator, geometry, and learning of dynamical systems},
  author={Mezi{\'c}, Igor},
  journal={Not. Am. Math. Soc},
  volume={68},
  number={7},
  pages={1087--1105},
  year={2021}
}

@article{Mezic2005,
  title={Spectral properties of dynamical systems, model reduction and decompositions},
  author={Mezi{\'c}, Igor},
  journal={Nonlinear Dyn.},
  volume={41},
  pages={309--325},
  year={2005},
  publisher={Springer}
}

@Article{Arbabi2017,
  author       = {Arbabi, Hassan and Mezić, Igor},
  date         = {2017-01},
  journaltitle = {SIAM J. Appl. Dyn. Syst.},
  title        = {Ergodic Theory, Dynamic Mode Decomposition, and Computation of Spectral Properties of the Koopman Operator},
  issn         = {1536-0040},
  number       = {4},
  pages        = {2096--2126},
  volume       = {16},
  publisher    = {Society for Industrial & Applied Mathematics (SIAM)},
}

@Article{Drmac2018,
  author       = {Zlatko Drma{\v{c}} and Igor Mezi{\'{c}} and Ryan Mohr},
  date         = {2018-01},
  journaltitle = {SIAM J. Sci. Comput.},
  title        = {Data Driven Modal Decompositions: Analysis and Enhancements},
  number       = {4},
  pages        = {A2253--A2285},
  volume       = {40},
  publisher    = {Society for Industrial {\&} Applied Mathematics ({SIAM})},
}

@Article{Drmac2020,
  author       = {Zlatko Drma{\v{c}} and Igor Mezi{\'{c}} and Ryan Mohr},
  date         = {2020-01},
  journaltitle = {SIAM J. Sci. Comput.},
  title        = {On Least Squares Problems with Certain Vandermonde--Khatri--Rao Structure with Applications to {DMD}},
  number       = {5},
  pages        = {A3250--A3284},
  volume       = {42},
  publisher    = {Society for Industrial {\&} Applied Mathematics ({SIAM})},
}

@Article{Cleveland1979,
  author       = {Cleveland, William S.},
  date         = {1979-12},
  journaltitle = {J. Am. Stat. Assoc.},
  title        = {Robust Locally Weighted Regression and Smoothing Scatterplots},
  issn         = {1537-274X},
  number       = {368},
  pages        = {829--836},
  volume       = {74},
  publisher    = {Informa UK Limited},
}

@Article{Frame2023,
  author       = {Peter Frame and Aaron Towne},
  date         = {2023-08},
  journaltitle = {{PLOS} {ONE}},
  title        = {Space-time {POD} and the Hankel matrix},
  editor       = {Mohamed Kamel Riahi},
  issn         = {1932-6203},
  number       = {8},
  pages        = {e0289637},
  volume       = {18},
  day          = {10},
  month        = {8},
  publisher    = {Public Library of Science ({PLoS})},
  year         = {2023},
}

@Article{Schmid2022,
  author       = {Peter J. Schmid},
  date         = {2022-01},
  journaltitle = {Annu. Rev. Fluid Mech.},
  title        = {Dynamic Mode Decomposition and Its Variants},
  number       = {1},
  pages        = {225--254},
  volume       = {54},
  publisher    = {Annual Reviews},
}

@Article{Bazina2024,
  author       = {Bazina, Tomislav and Mauša, Goran and Zelenika, Saša and Kamenar, Ervin},
  date         = {2024-05},
  journaltitle = {Robotics},
  title        = {Reducing Hand Kinematics by Introducing Grasp-Oriented Intra-Finger Dependencies},
  issn         = {2218-6581},
  number       = {6},
  pages        = {82},
  volume       = {13},
  publisher    = {MDPI AG},
}

@Article{Martinez2020,
  author       = {Itzel Jared Rodriguez Martinez and Andrea Mannini and Francesco Clemente and Christian Cipriani},
  date         = {2020-10},
  journaltitle = {IEEE Trans. Neural Syst. Rehabil. Eng.},
  title        = {Online Grasp Force Estimation From the Transient {EMG}},
  number       = {10},
  pages        = {2333--2341},
  volume       = {28},
  publisher    = {Institute of Electrical and Electronics Engineers ({IEEE})},
}

@Article{Martinez2020a,
  author       = {Itzel Jared Rodriguez Martinez and Andrea Mannini and Francesco Clemente and Angelo Maria Sabatini and Christian Cipriani},
  date         = {2020-02},
  journaltitle = {J. Neural Eng.},
  title        = {Grasp force estimation from the transient {EMG} using high-density surface recordings},
  number       = {1},
  pages        = {016052},
  volume       = {17},
  publisher    = {{IOP} Publishing},
}

@Article{Zhang2021,
  author       = {Na Zhang and Ke Li and Guanglin Li and Raviraj Nataraj and Na Wei},
  date         = {2021},
  journaltitle = {IEEE Trans. Neural Syst. Rehabil. Eng.},
  title        = {Multiplex Recurrence Network Analysis of Inter-Muscular Coordination During Sustained Grip and Pinch Contractions at Different Force Levels},
  pages        = {2055--2066},
  volume       = {29},
  publisher    = {Institute of Electrical and Electronics Engineers ({IEEE})},
}

@Article{Baranski2014,
  author       = {R. Bara{\'{n}}ski and A. Kozupa},
  date         = {2014-04},
  journaltitle = {Acta Phys. Pol. A},
  title        = {Hand Grip-{EMG} Muscle Response},
  number       = {4A},
  pages        = {A--7--A--10},
  volume       = {125},
  publisher    = {Institute of Physics, Polish Academy of Sciences},
}

@Article{Siavashani2023,
  author       = {A. Ghorbani Siavashani and A. Yousefi-Koma and A. Vedadi},
  date         = {2023-04},
  journaltitle = {J. Braz. Soc. Mech. Sci. Eng.},
  title        = {Estimation and early prediction of grip force based on {sEMG} signals and deep recurrent neural networks},
  number       = {5},
  volume       = {45},
  publisher    = {Springer Science and Business Media {LLC}},
}

@Article{Wu2021,
  author       = {Wu, Changcheng and Cao, Qingqing and Fei, Fei and Yang, Dehua and Xu, Baoguo and Zhang, Guanglie and Zeng, Hong and Song, Aiguo},
  date         = {2021-03},
  journaltitle = {PLOS ONE},
  title        = {Optimal strategy of sEMG feature and measurement position for grasp force estimation},
  editor       = {Wen, Li},
  issn         = {1932-6203},
  number       = {3},
  pages        = {e0247883},
  volume       = {16},
  publisher    = {Public Library of Science (PLoS)},
}

@Article{Khan2024,
  author       = {Khan, Salman Mohd and Khan, Abid Ali and Farooq, Omar},
  date         = {2024-04},
  journaltitle = {Heliyon},
  title        = {An early force prediction control scheme using multimodal sensing of electromyography and digit force signals},
  issn         = {2405-8440},
  number       = {8},
  pages        = {e28716},
  volume       = {10},
  publisher    = {Elsevier BV},
}

@Article{Stegeman2000,
  author       = {Stegeman, Dick F and Blok, Joleen H and Hermens, Hermie J and Roeleveld, Karin},
  date         = {2000-10},
  journaltitle = {J. Electromyogr. Kinesiol.},
  title        = {Surface EMG models: properties and applications},
  issn         = {1050-6411},
  number       = {5},
  pages        = {313--326},
  volume       = {10},
  publisher    = {Elsevier BV},
}

@Article{Ma2020,
  author       = {Ma, Ruyi and Zhang, Leilei and Li, Gongfa and Jiang, Du and Xu, Shuang and Chen, Disi},
  date         = {2020-06},
  journaltitle = {Alex. Eng. J.},
  title        = {Grasping force prediction based on sEMG signals},
  issn         = {1110-0168},
  number       = {3},
  pages        = {1135--1147},
  volume       = {59},
  publisher    = {Elsevier BV},
}

@Misc{Heckert2012,
  author    = {Heckert, N Alan and Filliben, James J and Croarkin, C M and Hembree, B and Guthrie, William F and Tobias, P and Prinz, J},
  date      = {2012},
  title     = {NIST/SEMATECH e-Handbook of Statistical Methods (NIST Handbook 151)},
  copyright = {License Information for NIST data},
  publisher = {National Institute of Standards and Technology},
}

@article{huo2023effects,
  title={Effects of EMG-based robot for upper extremity rehabilitation on post-stroke patients: a systematic review and meta-analysis},
  author={Huo, Yunxia and Wang, Xiaohan and Zhao, Weihua and Hu, Huijing and Li, Le},
  journal={Front. Physiol.},
  volume={14},
  pages={1172958},
  year={2023},
  publisher={Frontiers Media SA}
}

@Article{Miklashevsky2022,
  author       = {Miklashevsky, A.},
  date         = {2022-07},
  journaltitle = {PLOS ONE},
  title        = {Catch the star! Spatial information activates the manual motor system},
  editor       = {Myachykov, Andriy},
  issn         = {1932-6203},
  number       = {7},
  pages        = {e0262510},
  volume       = {17},
  publisher    = {Public Library of Science (PLoS)},
}

@article{singh2025koopman,
  author    = {Singh, Mayank and Hakam, Noor and Kesar, Trisha M and Sharma, Nitin},
  title     = {Koopman-Based Model Predictive Control of Functional Electrical Stimulation for Ankle Dorsiflexion and Plantarflexion Assistance},
  year      = {2025},
  volume    = {33},
  journal   = {IEEE Trans. Neural Syst. Rehabil. Eng.},
}

@article{qian2022deep,
  title={Deep multi-modal learning for joint linear representation of nonlinear dynamical systems},
  author={Qian, Shaodi and Chou, Chun-An and Li, Jr-Shin},
  journal={Sci. Rep.},
  volume={12},
  number={1},
  pages={12807},
  year={2022},
  publisher={Nature Publishing Group UK London}
}

@article{Li2014,
  author       = {Li, Huihui and Zhao, Guoru and Zhou, Yongjin and Chen, Xin and Ji, Zhen and Wang, Lei},
  date         = {2014-01},
  journaltitle = {BioMedical Engineering OnLine},
  title        = {Relationship of EMG/SMG features and muscle strength level: an exploratory study on tibialis anterior muscles during plantar-flexion among hemiplegia patients},
  issn         = {1475-925X},
  number       = {1},
  volume       = {13},
  publisher    = {Springer Science and Business Media LLC},
}

\end{document}